\documentclass[conference]{IEEEtran}
\IEEEoverridecommandlockouts 

\usepackage[T1]{fontenc}
\usepackage{amsmath}
\usepackage{newtxtext,newtxmath} 
\usepackage{microtype}

\usepackage{graphicx}
\usepackage{xcolor}
\usepackage{booktabs}
\usepackage{array}
\usepackage{tabularx}
\usepackage{makecell}
\usepackage{pgfplots}   
\pgfplotsset{compat=1.18}
\usetikzlibrary{calc,arrows.meta}
\usepackage{cite}
\usepackage[colorlinks=true,linkcolor=linkblue,citecolor=linkblue,urlcolor=linkblue,
  pdftitle={Multi-Speed Field Evaluation of a Semi-Autonomous Precision Sprayer for Weed Control in Vegetable Crops},
  pdfauthor={Md Khairul Islam; Boyang Deng; Yuzhen Lu; Daniel Brainard; Ajit Srivastava},
  pdfsubject={Precision spraying; weed detection; speed-accuracy trade-off},
  pdfkeywords={precision agriculture; autonomous systems; smart sprayer; weed management; vegetable production}]{hyperref}
\definecolor{linkblue}{RGB}{20,60,140}

\makeatletter
\g@addto@macro\UrlBreaks{\do\-\do\=\do\&}
\makeatother

\definecolor{ColW}{RGB}{31,119,180}
\definecolor{ColC}{RGB}{192,32,32}
\definecolor{ColM}{RGB}{112,48,160}
\definecolor{ColG}{RGB}{30,120,60}

\newcommand{\WHR}{\mathrm{WHR}}
\newcommand{\CCR}{\mathrm{CCR}}
\newcommand{\IoU}{\mathrm{IoU}}
\newcommand{\mAP}{\mathrm{mAP}}
\newcommand{\tabnote}[1]{\par\vspace{3pt}{\footnotesize\raggedright #1\par}}
\tikzset{ptag/.style={anchor=north west,fill=white,fill opacity=0.85,text opacity=1,
  inner sep=1.5pt,font=\footnotesize\bfseries}}
\newcommand{\panelimg}[3]{%
  \begin{tikzpicture}[baseline=(img.center),inner sep=0pt]
    \node (img) {\includegraphics[#1]{#2}};
    \node[ptag] at ([shift={(2pt,-2pt)}]img.north west) {#3};
  \end{tikzpicture}}

\newcommand{\Vijay}{\mbox{Vi\kern0pt jayakumar}}
\IEEEaftertitletext{\vspace{-0.45cm}}

\begin{document}

\title{\fontsize{21}{25}\selectfont Multi-Speed Field Evaluation of a Semi-Autonomous Precision Sprayer for Weed Control in Vegetable Crops}

\author{
\IEEEauthorblockN{Md Khairul Islam\textsuperscript{1}, Boyang Deng\textsuperscript{1}, Yuzhen Lu\textsuperscript{1,*}, Daniel Brainard\textsuperscript{2}, and Ajit Srivastava\textsuperscript{1}}
\IEEEauthorblockA{\textsuperscript{1}Department of Biosystems and Agricultural Engineering, \textsuperscript{2}Department of Horticulture\\
Michigan State University, East Lansing, MI 48824, USA}
\thanks{*Corresponding author: Yuzhen Lu (luyuzhen@msu.edu).}
}

\maketitle
\thispagestyle{plain}

\begin{abstract}
\mdseries
Machine vision-guided precision spraying is promising for sustainable vegetable production by enabling site-specific weed targeting, reducing herbicide application while protecting crop yield and quality. However, the effects of operating speed of sprayer systems on the integrated detection--tracking--actuation pipeline remain insufficiently characterized under field conditions. This study evaluated a semi-autonomous, tractor-based spraying system for targeted weed control in radish and quantified performance across three operating speeds. A five-season dataset comprising 10,599 images and 84,311 annotated instances across seven classes was curated for model development. A YOLOv11-medium model was integrated with ByteTrack tracking and a customized 12-nozzle sprayer on a semi-autonomous electric tractor. Field trials were conducted at 0.13, 0.27, and 0.45~m/s (i.e., 0.3, 0.6, and 1.0~mph) to characterize speed-dependent changes in weed detection, targeting, actuation, and crop damage. At the lowest speed of 0.13~m/s, the system achieved 94.6\% mAP@50 and 93.2\% recall for weed and crop detection. However, increasing platform speed substantially reduced system performance, with weed hit rate declining from 78.3\% to 35.0\% and crop contact rates increasing from 20.1\% to 28.7\%. Consequently, the system did not meet predefined commercial-viability thresholds. These results demonstrate that practical deployment requires improvements beyond weed detection alone, particularly coordinated improvements in motion-robust plant perception, spray-timing compensation, and crop-adjacent actuation precision. The identified performance bottlenecks provide specific engineering targets for advancing vision-guided precision spraying toward practical, high-speed vegetable weed management. The software programs developed in this study have been made publicly available to support future precision weeding research.
\end{abstract}

\begin{IEEEkeywords}
\mdseries
Precision agriculture, Autonomous systems, Smart sprayer, Weed detection, YOLOv11
\end{IEEEkeywords}

\section{Introduction}
\label{sec:intro}
Weeds represent one of the most economically damaging threats in global crop production, causing an estimated 34\% reduction in attainable yield across major food crops worldwide~\cite{oerke2006}, with the problem compounded by rapidly growing herbicide resistance now documented in 275 weed species across 76 countries~\cite{heap2026}. Three decades of stagnation in new herbicide mode-of-action discovery~\cite{westwood2018} underscores the urgent need for precision weed control strategies that reduce chemical dependency while maintaining production efficiency. Vegetable production systems face particular vulnerability. Weak crop competitive ability, limited registered herbicide options, and critical timing requirements create severe management challenges~\cite{bell2022}. In brassica crops, including radish and cabbage, weed competition during critical periods causes 52\nobreakdash--92\% yield reductions~\cite{miller1991,sutharsan2015}. Recent mechanistic research reveals that weeds reduce yields not merely through resource competition but through early signaling mechanisms involving light quality and volatile compounds, rendering yield loss largely irreversible after narrowing developmental windows~\cite{horvath2023}. Combined with narrow profit margins and limited chemical options, vegetable systems are ideal candidates for precision weeding technologies, achieving site-specific control with minimal crop injury.

Machine-vision-guided precision spraying offers pathways to dramatically reduce herbicide inputs while maintaining efficacy. Field demonstrations document herbicide reductions of 23\nobreakdash--89\% with cost savings reaching 56\% compared to broadcast application~\cite{timmermann2003,gerhards2022}. Deep learning, particularly YOLO-family detectors, has transformed weed detection through robust feature extraction and multi-class recognition across diverse conditions~\cite{chen2022, DengLu2026}. Recent research prototypes demonstrate promising capabilities: AI-powered sprayers for tomato and pepper achieved 88\nobreakdash--89\% precision with $<$5\% missed targets~\cite{vijayakumar2025}, UAV-based spot spraying in maize achieved 94\% classification accuracy with 47\% herbicide savings at 6~km/h~\cite{allmendinger2024}, and large-scale sugarcane validation demonstrated 97\% control efficacy with 35\% herbicide reduction~\cite{azghadi2025}. Our own group's prior prototypes illustrate this iterative development: an initial 12-nozzle ATV-mounted sprayer using YOLOv10 achieved a video-based mAP@50 of 48.9\% and a 51.3\% spraying hit rate under field conditions~\cite{deng2026a}, while a subsequent 24-nozzle, YOLOv11-based upgrade improved field performance to 69.3\% mAP@50 and a 74.6\% spraying hit rate in a radish field, though crop contact remained substantial~\cite{deng2026c}.

Although substantial progress has been made, current literature lacks a systematic characterization of speed--accuracy trade-offs, specifically varying vehicle speed across controlled levels while concurrently measuring detection accuracy, actuation precision, and weed control efficacy at each speed~\cite{gerhards2024}. A further gap concerns accessibility; most reported systems rely on custom research-grade hardware, whereas low-cost designs retrofittable to existing tractors, critical for adoption by small and mid-sized vegetable producers, remain comparatively rare~\cite{terra2021}. Previous studies rely on single operational speeds or even static image assessments, neglecting performance degradation caused by motion blur and reduced frame overlap during continuous video processing~\cite{du2024}. Notably, multi-season model robustness remains inadequately characterized, with cross-season testing revealing accuracy degradation up to 14.5\%~\cite{deng2024}. Dedicated multi-season data collection efforts remain limited. Recent multi-temporal datasets spanning multiple imaging campaigns have been developed for orchard weed detection~\cite{salcedo2025}, but comparable multi-season resources for vegetable row-crop systems, spanning successive growing seasons as pursued in objective~(i) of this study, are largely absent from the published literature. Furthermore, in-field crop damage rates are rarely reported specifically for vision-guided precision spraying systems~\cite{gerhards2024}; by contrast, the mechanical weed control literature has documented crop damage extensively, from quantified stand reductions under intelligent cultivators~\cite{lati2016} to detailed intra-row yield-loss curves spanning a wide range of weed densities~\cite{melander2021}.

To address these gaps, the study executes a rigorous field-based assessment of a semi-autonomous precision spraying apparatus, prioritizing the quantification of trade-offs between speed and accuracy in production field environments. The methodological framework synergizes a YOLOv11 detection model optimized on a comprehensive five-season dataset (10,599 images, 84,311 instances, 7 target classes), dynamic trajectory tracking via ByteTrack, and a customized 12-nozzle actuation matrix, fully integrated onto an autonomous tractor guided by RTK-GPS technology. The system was evaluated through systematic multi-speed field trials at 0.13, 0.27, and 0.45~m/s (0.3, 0.6, and 1.0~mph), with comprehensive performance assessment, including crop damage quantification. The primary objectives are as follows:
\begingroup
\renewcommand{\theenumi}{\roman{enumi}}
\renewcommand{\labelenumi}{(\theenumi)}
\begin{enumerate}
    \item A five-season weed and crop dataset was developed to support robust model training across diverse field conditions, and a YOLOv11-medium detection model integrated with ByteTrack multi-object tracking was trained and evaluated for real-time weed detection on a tractor-based semi-autonomous spraying platform.
    \item Systematic multi-speed field trials at three operating speeds (0.13, 0.27, and 0.45~m/s) were conducted to characterize speed-dependent degradation across the full detection--tracking--actuation pipeline, including crop damage quantification.
    \item Key technical limitations, including detection recall degradation under motion, spray-timing latency, and crop-adjacent targeting precision, were identified, and a prioritized improvement roadmap addressing these bottlenecks was established toward achieving commercial viability thresholds.
\end{enumerate}
\endgroup

\section{Materials and Methods}
\label{sec:methods}

\subsection{System Architecture and Hardware}
\label{sec:arch}
\begin{figure*}[!t]
    \centering
    \panelimg{trim=424bp 1078bp 375bp 28bp,clip,height=1.50in}{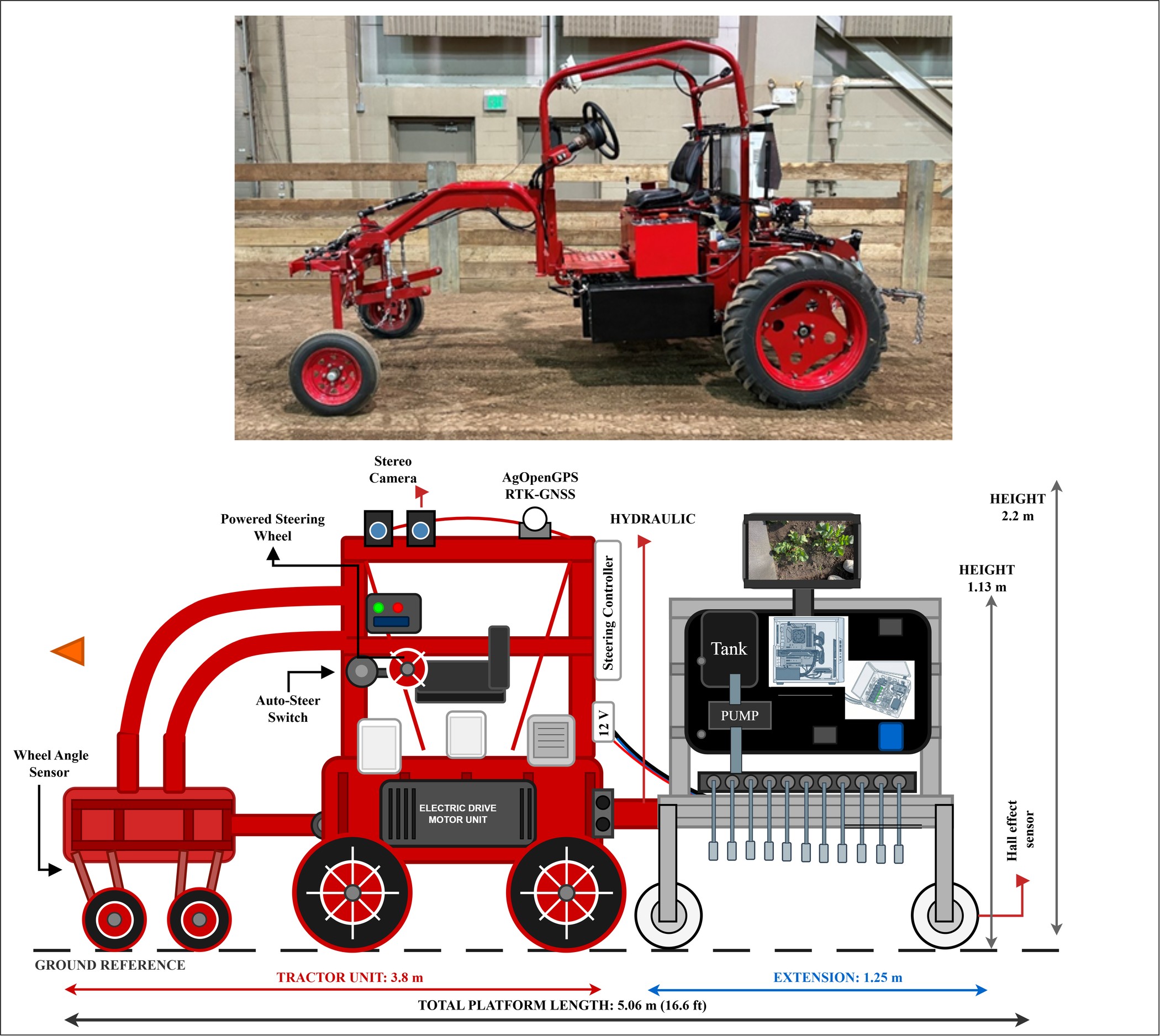}{(a)}\hfill
    \panelimg{trim=17bp 9bp 168bp 817bp,clip,height=2.45in}{images/fig1_platform.jpg}{(b)}
    \caption{Semi-autonomous precision spraying platform: (a) photograph of the converted electric tractor (Oggún II--based; Section~\ref{sec:platform}) and (b) schematic showing navigation, imaging with onboard computing, and the sprayer actuator. Total platform length: 5.06~m (16.6~ft), comprising a 3.81~m tractor unit and a 1.25~m spray extension.}
    \label{fig:platform}
\end{figure*}

\begin{figure*}[!t]
    \centering
    \includegraphics[width=0.94\textwidth]{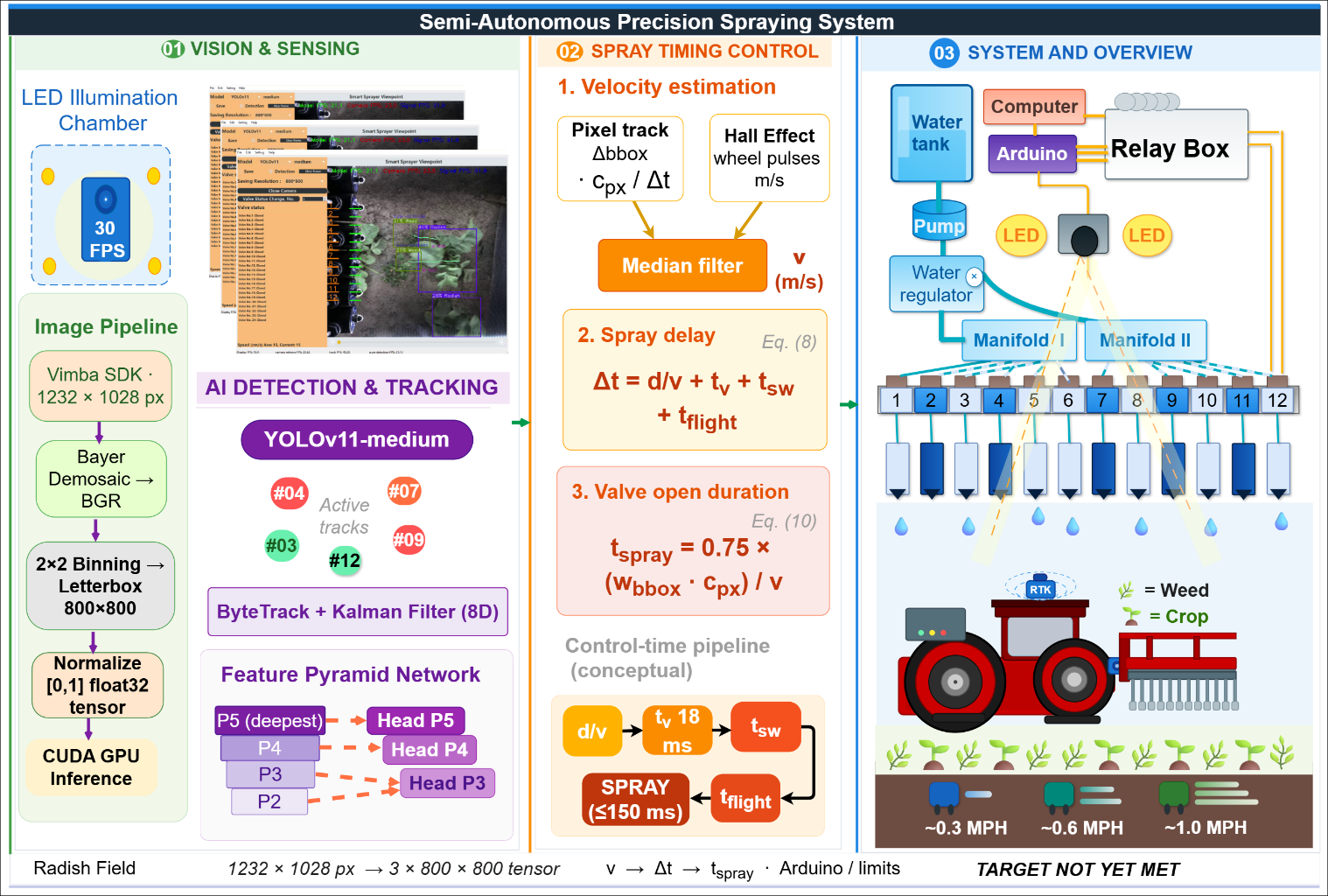}
    \caption{System architecture, organized into three functional subsystems. (1) Vision and sensing: an enclosed imaging chamber acquires frames at 30~frames per second, which are preprocessed and passed to the YOLOv11-medium detector with ByteTrack tracking. (2) Spray timing control: platform velocity is estimated from Hall Effect wheel pulses and pixel-based tracking, median filtered, and used to compute the spray delay (Eq.~(\ref{eq:delay})) and valve open duration (Eq.~(\ref{eq:spray})) for each nozzle. (3) Hardware integration: onboard computer, Arduino relay box, pump, and dual manifolds driving 12 independently actuated nozzles at three operating speeds (0.3--1.0~mph). The control-time pipeline enforces a $\leq$150~ms total response window.}
    \label{fig:architecture}
\end{figure*}

Fig.~\ref{fig:platform} shows the converted electric tractor platform (Section~\ref{sec:platform}) that served as the base for the precision spraying system described in this section. The precision spraying system integrates YOLOv11-based weed detection, ByteTrack multi-object tracking, and synchronized spray actuation on a semi-autonomous electric tractor platform. The modular architecture employs concurrent processing threads for camera acquisition, detection inference, and valve control, with inter-thread communication via queue-based data structures ensuring real-time coordination without memory accumulation. As shown in Fig.~\ref{fig:architecture}, the system operates as three coordinated functional subsystems (vision and sensing, spray timing control, and hardware integration), each detailed in the subsections below.

The hardware comprises four integrated subsystems: (1) autonomous navigation platform with RTK-GPS guidance and auto-steering, (2) enclosed imaging chamber with LED illumination and industrial camera, (3) onboard computing unit (Intel i9-11900K, NVIDIA RTX 4060 Ti), and (4) spray actuator assembly with 12 independently controlled nozzles. An extensive configuration system manages hardware calibration, detection thresholds, timing offsets, and filtering parameters, enabling rapid tuning without code modification.

\subsection{Autonomous Platform and Navigation}
\label{sec:platform}
The system was deployed on a converted Oggún II tractor (Ronnie Baugh Tractors) (Fig.~\ref{fig:platform}a), a 1,600~lb (726~kg) compact platform assembled from standard off-the-shelf parts, with stock specifications of 150~in.\ (3.81~m) length, 24~in.\ ground clearance, independent hydraulic drive, hydraulic steering and brakes, zero-turn capability, a differential lock, and a Category 1 three-point hitch; front track width is adjustable from 52 to 72~in.\ in 2~in.\ increments, with a continuously variable rear track over the same range~\cite{darrow2026}.

For this study, the platform's original 22~HP Honda twin-cylinder gasoline engine was replaced with a 37~HP (27.6~kW) three-phase AC brushless induction motor, a 68\% increase in available power over the stock gasoline configuration, drawing from four 88~V (76.8~V nominal) Smartec LiFePO$_4$ battery packs wired in parallel (56~kg each, $\sim$110~Wh/kg, 24.6~kWh total capacity). The electric motor directly drives a variable-displacement axial-piston traction pump, controlled via a mechanical foot pedal, which supplies hydraulic flow to two fixed-displacement orbital wheel motors on the rear wheels; an auxiliary gear pump mounted on the motor shaft maintains continuous hydraulic flow to the power steering and implement lift cylinders~\cite{darrow2026}.

This platform is distinct from the Farm-Ng Amiga-based prototypes used in our group's prior work~\cite{deng2026a,deng2026c}: the Oggún II's larger hydrostatic-drive chassis and three-point hitch provide the load capacity and stability needed for the heavier onboard spray system and extended 1.25~m spray extension used in this study, at the cost of a larger minimum turning footprint relative to the compact Amiga platform. Throttle control remained under direct manual operator supervision, via the foot pedal, to isolate spraying performance from autonomous path planning, while auto-steering maintained precise row following.

Electrical power was distributed across two independent systems. Each Smartec battery pack incorporates an internal DC--DC converter regulated to a 12~V, 15~A output, which supplied the platform's legacy 12~V electrical architecture (key switch, motor management module, CAN-bus termination) as well as auxiliary components including the electric spray pump, solenoid valves, and LED illumination. A separate ECO-WORTHY 12.8~V, 150~Ah LiFePO$_4$ battery (1.92~kWh nominal capacity), housed in a dedicated enclosure with an integrated 120~A battery management system, provided independent power for the onboard computing unit and camera system, isolating the imaging and detection pipeline from traction-system load transients.

Navigation employed an open-source auto-steering system based on AgOpenGPS~\cite{tischler2024} with an on-farm RTK-GPS base station providing differential corrections to a rover-mounted multi-band GNSS receiver via radio link. The auto-steering assembly integrated a powered steering motor, wheel angle sensor, multi-band GNSS receiver, and steering controller, coordinating sensor inputs to generate motor commands. Total navigation retrofit costs approximately \$2,150 (Supplementary Table S1), substantially less than commercial systems while achieving comparable performance.

\subsection{Vision and Computing System}
\label{sec:vision}
Image acquisition employed an industrial color camera (Alvium G5-508c, Allied Vision Technologies, Stadtroda, Germany) fitted with a 5~mm focal length lens (Kowa, Japan). The camera, which features a 5.1~MP CMOS sensor, was operated with $2\times 2$-pixel binning, giving an effective resolution of $1{,}232\times1{,}028$ pixels to balance spatial resolution with processing throughput. The camera was interfaced via Gigabit Ethernet at 30~FPS, with the exposure time fixed at $\sim$14~ms to prevent motion blur while maximizing brightness under LED illumination. This fixed exposure establishes a fundamental physical constraint on motion-induced image blur. The displacement of a scene element during a single exposure is
\begin{equation}
    \delta_{\mathrm{blur}} = v\,t_{\mathrm{exp}},
    \label{eq:blur}
\end{equation}
where $v$ is the platform forward speed and $t_{\mathrm{exp}} \approx 14$~ms is the exposure time. Expressed in pixel units using the calibrated ground-sampling distance $c_{\mathrm{px}} = 0.838$~mm/pixel,
\begin{equation}
    \delta_{\mathrm{px}} = \frac{v\,t_{\mathrm{exp}}}{c_{\mathrm{px}}}.
    \label{eq:blurpx}
\end{equation}
At the platform forward speed of 0.13~m/s (0.3~mph), $\delta_{\mathrm{px}} \approx 2.2$~pixels; at 0.45~m/s (1.0~mph) this increases to $\delta_{\mathrm{px}} \approx 7.5$~pixels, a 3.4-fold increase that renders sub-centimeter morphological features of small weeds ($<$2.5~cm) indistinguishable from background texture, providing the physical basis for the speed-dependent recall degradation characterized in Section~\ref{sec:results}.

The camera was mounted vertically 0.8~m above the ground within an enclosed imaging chamber. Consistent illumination was provided by white LED arrays within the enclosed chamber, which served three critical functions: eliminating sunlight variability and specular reflections, maintaining consistent shadow patterns regardless of ambient conditions, and extending the operating window to low-light conditions. The lightweight aluminum frame with opaque fabric panels excluded ambient light while allowing maintenance access.

Real-time detection was executed on vehicle-mounted computing (Intel i9-11900K, 64~GB RAM, NVIDIA RTX 4060 Ti GPU with 16~GB VRAM, Ubuntu 22.04, CUDA 12.8). Actual detection throughput ranged from 15\nobreakdash--30~FPS, depending on scene complexity. The spatial frame density (detection opportunities per meter of travel) is
\begin{equation}
    \rho_{f} = \frac{f_{\mathrm{cam}}}{v},
    \label{eq:density}
\end{equation}
where $f_{\mathrm{cam}} = 30$~FPS is the camera frame rate. At 0.13~m/s (0.3~mph), $\rho_{f} \approx 230$~frames/m; at 0.45~m/s (1.0~mph) this falls to 67~frames/m, a 71\% reduction in per-plant detection opportunities that directly compounds the blur degradation described in Eqs.~(\ref{eq:blur})--(\ref{eq:blurpx}). Complete hardware specifications appear in Supplementary Table S1.

\subsection{Spray Actuator Design}
\label{sec:actuator}
\begin{figure*}[!t]
    \centering
    \panelimg{width=0.78\textwidth}{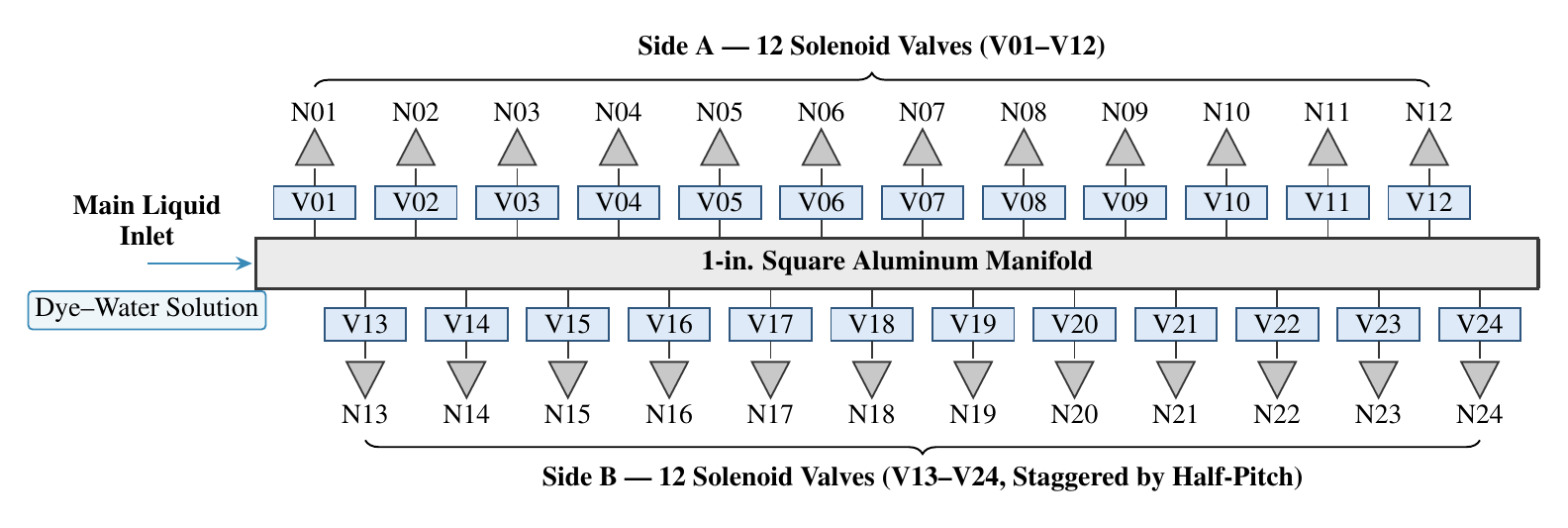}{(a)}\\[4pt]
    \panelimg{height=2.05in}{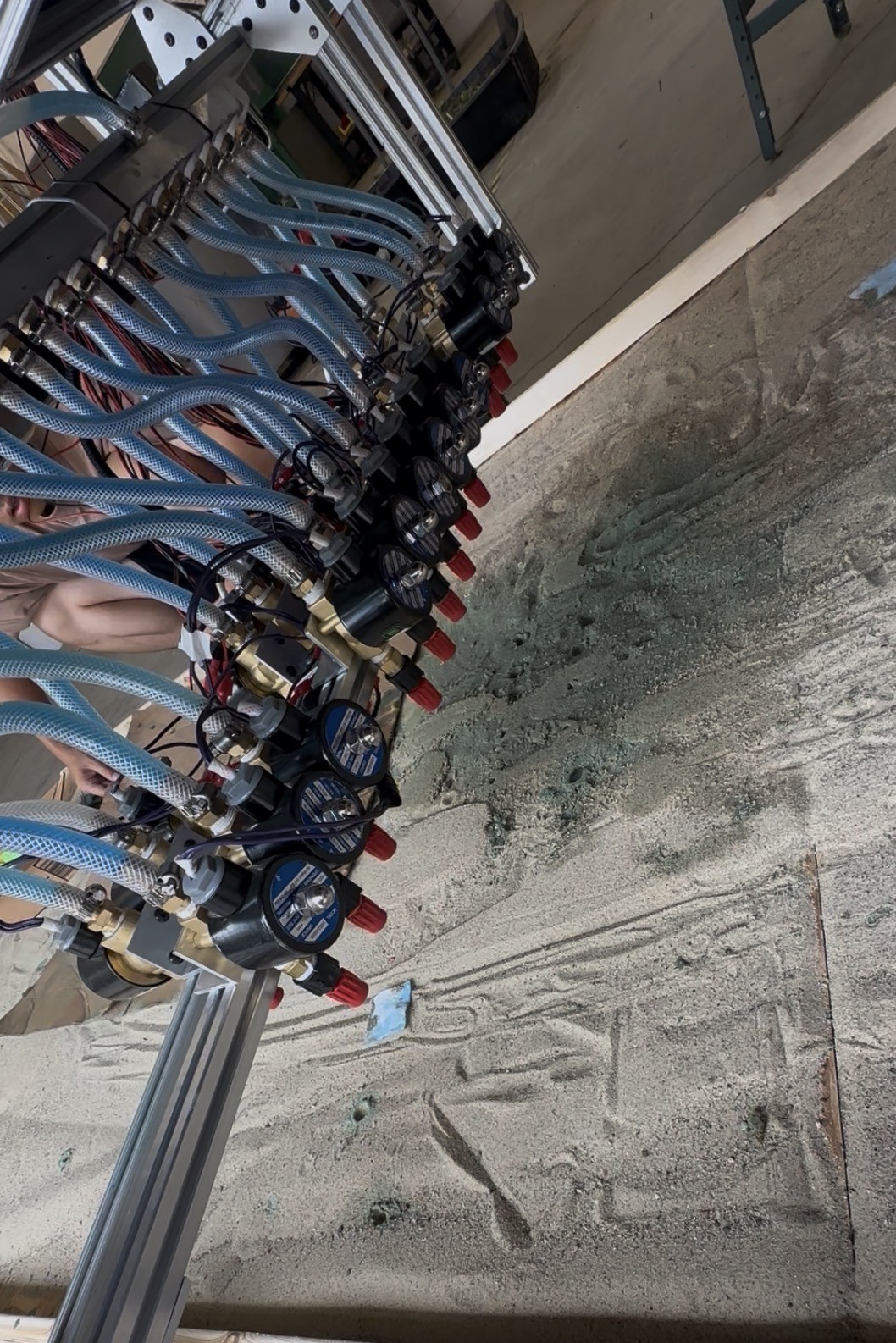}{(b)}\hspace{6pt}%
    \panelimg{height=2.05in}{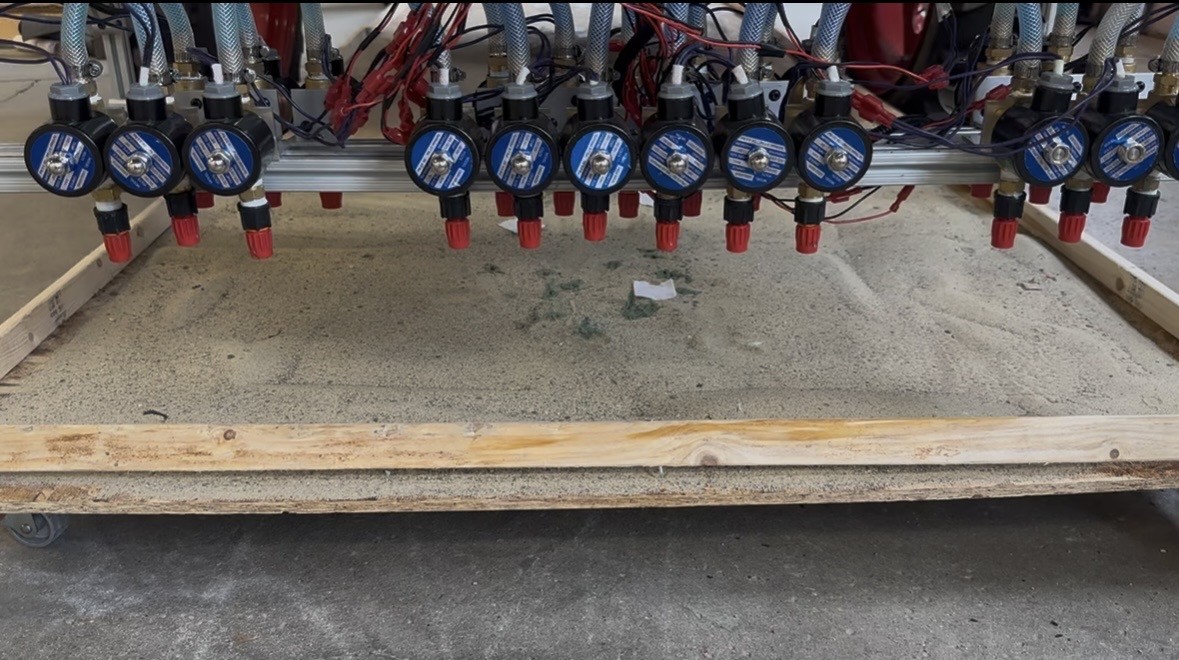}{(c)}
    \caption{Nozzle--solenoid valve--manifold assembly used in this study: (a) schematic and (b), (c) photographs.}
    \label{fig:nozzle}
\end{figure*}

The spray actuator comprised 12 independently controlled nozzles (Figs.~\ref{fig:architecture}--\ref{fig:nozzle}) in a linear array at 10~mm center-to-center spacing (total coverage width $\sim$130~mm). This narrow spacing prioritized maximum spatial resolution for targeting accuracy characterization, accepting reduced total coverage as a trade-off.

Each nozzle assembly used modified check valves producing concentrated spray patterns ($\sim$15~mm diameter at soil surface). Spray fluid (18.9~L polyethylene tank) was pressurized by a 12~V DC diaphragm pump with pressure regulation at 20~psi, balancing spray pattern formation against drift. Custom manifolds, constructed from $1\times1$~in.\ square aluminum tube with 3D-printed end caps sealed using thread-sealing tape and adhesive, positioned nozzles perpendicular to the soil surface. Spray actuation employed normally closed solenoid valves (USS2-00006, U.S. Solid, Cleveland, OH; manufacturer-rated response time $\leq$1~s, a conservative full-stroke rating typical of general-purpose valves of this type) controlled via an Arduino microcontroller with automatic safety closure (150~ms timeout) preventing uncontrolled spraying from communication failures. The spray-timing model (Section~\ref{sec:timing}) instead assumes a faster $\sim$18~ms nominal response, consistent with the partial-stroke, low-pressure actuation used here rather than the manufacturer's full-stroke rating; this assumption was not independently bench-validated on this hardware (Section~\ref{sec:cropdamage}). Platform velocity was measured via a Hall Effect wheel sensor, supplemented by pixel-based estimation from tracked plant (weed) movement, with median filtering across sources providing robust speed estimation.

\subsection{Dataset Development}
\label{sec:dataset}
The detection model was trained on a five-season dataset comprising three sub-datasets acquired over five seasons, i.e., 3SeasonWeedDet10, Weed\&CropDet2024, and Weed\&CropDet2025. The combined dataset contained 10,599 annotated images containing 84,311 object instances across seven target classes (Table~\ref{tab:dataset}). This multi-season approach addressed temporal generalization challenges where single-season models exhibit up to 14.5\% accuracy degradation in subsequent seasons~\cite{deng2024}.

3SeasonWeedDet10 is a previously published dataset~\cite{denglu2025}. Weed\&CropDet2024 comprises 2024 field weed imagery from Michigan State University, reused here with its original weed-only annotations rather than re-annotated for this study. The Weed\&CropDet2025 component (1,161 images, 19,338 instances), collected June--August 2025 using the tractor-mounted platform under deployment conditions, provides training data directly representative of field trial imaging geometry; a related radish precision-sprayer platform and dataset from the same research group are described in Deng \textit{et al.}~\cite{deng2026c}, and this study's distinct contribution lies in the multi-speed field characterization enabled by this imagery (Sections~\ref{sec:expval} and~\ref{sec:perfeval}) rather than in the underlying imagery being wholly exclusive to this work. For field trials, the dataset was consolidated into seven classes reflecting experimental site composition (Table~\ref{tab:classes}).

\begin{table*}[!t]
\centering
\caption{Dataset Composition Across Five Growing Seasons}
\label{tab:dataset}
\footnotesize
\renewcommand{\arraystretch}{1.12}
\begin{tabularx}{\textwidth}{@{}lrrllX@{}}
\toprule
Dataset & Images & Instances & Species/classes & Collection period & Primary imaging platform \\
\midrule
3SeasonWeedDet10\textsuperscript{a} & 8,436 & 27,963 & 10 weed species & 2021\nobreakdash--2023 & Smartphones, digital cameras, vehicle-mounted \\
Weed\&CropDet2024 & 1,002 & 37,010 & 5 crops, 12 weeds & 2024 & Vehicle-mounted platform \\
Weed\&CropDet2025 & 1,161 & 19,338 & Radish, 6 weeds & June--August 2025 & Tractor-mounted system (this study) \\
\midrule
\textbf{Combined total} & \textbf{10,599} & \textbf{84,311} & --- & 2021\nobreakdash--2025 & Multiple platforms \\
\bottomrule
\end{tabularx}
\tabnote{\textsuperscript{a}Previously published~\cite{denglu2025}.}
\end{table*}

For the newly collected Weed\&CropDet2025 component, bounding box annotations were manually created using LabelImg~\cite{tzutalin2015}, with each plant instance enclosed by the tightest axis-aligned rectangle and labeled according to the seven target classes (Table~\ref{tab:classes}); the previously published 3SeasonWeedDet10 and Weed\&CropDet2024 components retained their original published annotations. All annotations followed YOLO format with normalized center coordinates, width, and height values. Quality control on the newly annotated data employed dual-annotator review of 15\% of images with $>$90\% inter-annotator agreement (intersection over union, IoU $\geq 0.5$). The combined dataset was partitioned into training (75\%), validation (5\%), and test (20\%) subsets using stratified sampling at the field-plot level (preventing data leakage from temporally adjacent frames).

\begin{table*}[!t]
\centering
\caption{Target Species and Classes in the Combined Dataset}
\label{tab:classes}
\footnotesize
\renewcommand{\arraystretch}{1.12}
\begin{tabularx}{\textwidth}{@{}llcX@{}}
\toprule
Class name & Scientific name & Role & Notes \\
\midrule
Radish & \textit{Raphanus sativus} L. & Crop & Primary crop requiring herbicide protection \\
Carpetweed & \textit{Mollugo verticillata} L. & Weed & Summer annual broadleaf \\
Lambsquarters & \textit{Chenopodium album} L. & Weed & Summer annual broadleaf \\
Palmer amaranth & \textit{Amaranthus palmeri} S. Watson & Weed & Summer annual broadleaf \\
Smartweed & \textit{Persicaria} spp. & Weed & Summer annual broadleaf \\
Purslane & \textit{Portulaca oleracea} L. & Weed & Summer annual broadleaf \\
Weed (general) & Multiple species & Weed & Additional species and uncertain identifications \\
\bottomrule
\end{tabularx}
\end{table*}

Input images were preprocessed to $800\times800$ pixel resolution, preserving aspect ratio, with neutral gray padding (RGB: 128, 128, 128). Data augmentation was applied on-the-fly to training images only (validation and test images were evaluated unaugmented) using the Ultralytics default pipeline: mosaic augmentation, random flips, hue-saturation-value (HSV) perturbations (hue $\pm$0.015, saturation $\pm$0.7, value $\pm$0.4), scaling ($\pm$50\%), and translation ($\pm$10\%).

\subsection{Detection and Tracking System}
\label{sec:detTrack}

\subsubsection{Detection Model}
\label{sec:detmodel}
\begin{figure*}[!t]
    \centering
    \includegraphics[width=0.96\textwidth]{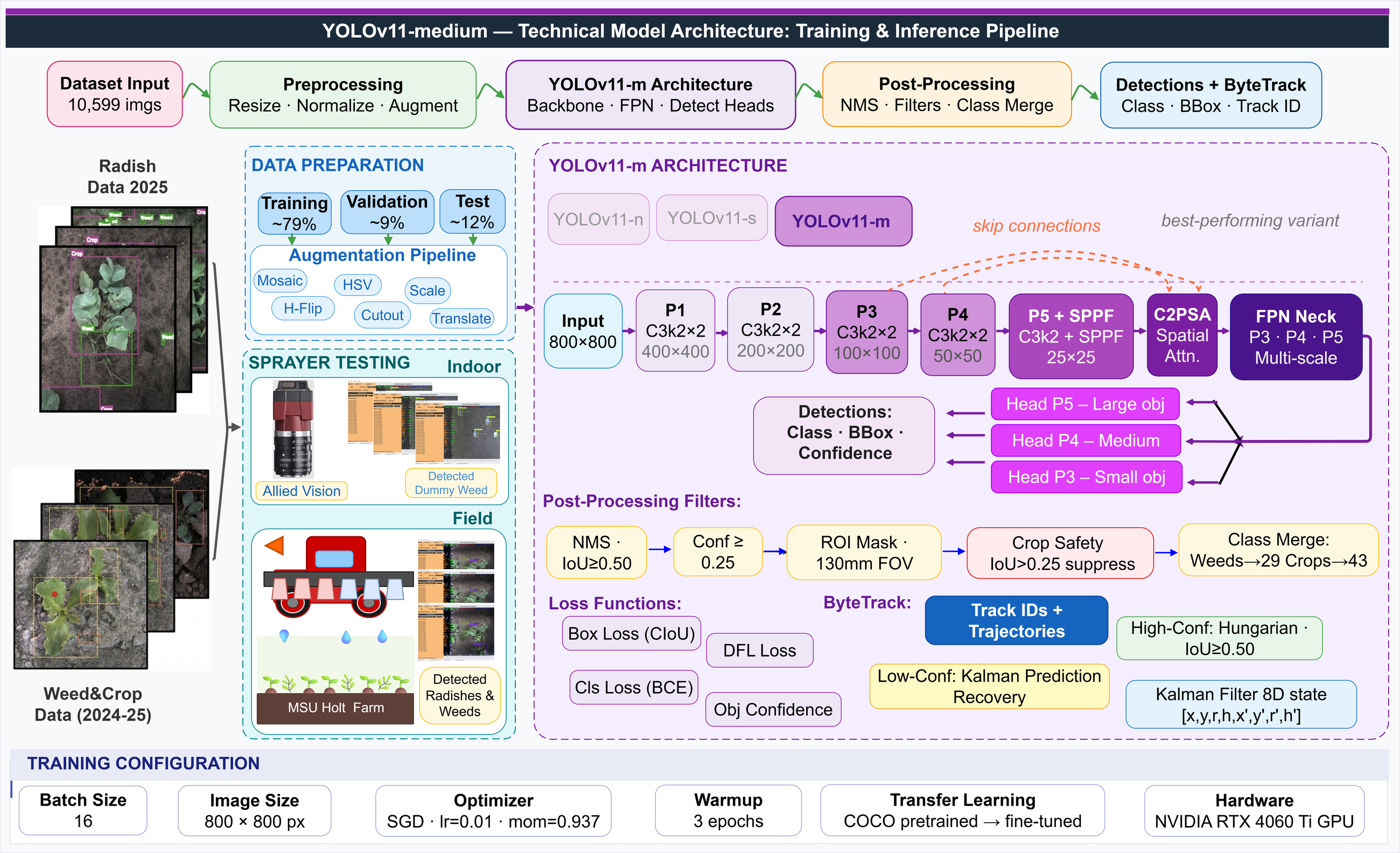}
    \caption{Overview of the YOLOv11 training pipeline and inference architecture. Three model variants (nano, small, and medium) were compared on the five-season dataset; YOLOv11-medium was selected for field deployment based on its superior mAP@50 (89.9\%) and real-time compatibility at 30~FPS. At inference, the model applies post-processing filters, removes duplicate detections, and tracks objects with ByteTrack.}
    \label{fig:yolo}
\end{figure*}

Weed detection employed YOLOv11-medium~\cite{jocher2024}, selected for optimal speed--accuracy balance in real-time agricultural applications; the complete training and inference pipeline is summarized in Fig.~\ref{fig:yolo}. In the broader weed-detection literature, YOLOv11 has achieved 13.5~ms inference with mAP@50 of 89.0\nobreakdash--92.0\% across benchmarked weed-detection tasks~\cite{sharma2024}, motivating its selection here for sustained real-time operation without detection latency confounding spray timing. The YOLOv11 architecture incorporates optimized Cross-Stage Partial (C3k2) backbone blocks and spatial attention modules (C2PSA) in the neck, enhancing small object detection critical for early-stage weeds. The detection head generates predictions at three scales, enabling simultaneous detection of small seedlings ($<$2~cm) and established plants ($>$10~cm).

Model training utilized the Ultralytics PyTorch implementation (version 11.0.0~\cite{jocher2024}; \url{https://github.com/ultralytics/ultralytics}), with 36 epochs and a batch size of 8 at $800\times800$ pixel resolution. Optimization employed stochastic gradient descent with momentum ($\beta = 0.937$), initial learning rate $\alpha_0 = 0.01$ with the standard cosine annealing schedule of Loshchilov and Hutter~\cite{loshchilov2017}, adopted without modification:
\begin{equation}
    \alpha_{t} = \alpha_{\min} + \frac{1}{2}\left(\alpha_{0} - \alpha_{\min}\right)\left[1 + \cos\!\left(\frac{\pi t}{T}\right)\right],
    \label{eq:cosine}
\end{equation}
where $t$ is the current epoch, $T = 36$ is the total number of epochs, and $\alpha_{\min} = 0.0001$. Weight decay $\lambda = 0.0005$ provided regularization. Complete hyperparameters are summarized in Supplementary Table S2. For field deployment, detection thresholds were confidence $\geq 0.25$ and IoU $\geq 0.5$ for non-maximum suppression.

\subsubsection{Multi-Object Tracking}
\label{sec:tracking}
Because a given plant's frame-level detection can be inconsistent across consecutive frames under motion blur and partial occlusion, spray triggering cannot rely on single-frame detections alone; multi-object tracking was used to maintain persistent plant identities across frames and supply the stabilized position and velocity estimates required by a spray-timing model (Section~\ref{sec:timing}). Multi-object tracking employed ByteTrack~\cite{zhang2022}, which maintains persistent object identities across frames through two-stage association: high-confidence matching (confidence $> 0.25$) using IoU-based Hungarian assignment~\cite{bewley2016}, followed by low-confidence recovery ($0.1 <$ confidence $< 0.25$) reducing track fragmentation from temporary detection failures. State estimation for each track employed a Kalman filter with an eight-dimensional state vector representing center position ($x$, $y$), aspect ratio $r$, height $h$, and their respective velocities, following the standard linear constant-velocity motion model:
\begin{equation}
\begin{aligned}
\mathbf{x}_{t} &= \mathbf{F}\,\mathbf{x}_{t-1} + \mathbf{w}_{t},\\
\mathbf{z}_{t} &= \mathbf{H}\,\mathbf{x}_{t} + \boldsymbol{\nu}_{t},
\end{aligned}
\label{eq:kalman}
\end{equation}
where $\mathbf{F}$ is the $8\times8$ state transition matrix implementing constant-velocity dynamics (position, aspect ratio, and height each updated by their respective velocity terms over the frame interval), $\mathbf{H}$ is the $4\times8$ observation matrix extracting the directly observed quantities ($x$, $y$, $r$, $h$) from the full state vector, and $\mathbf{w}_{t}$, $\boldsymbol{\nu}_{t}$ represent process and measurement noise with empirically tuned covariances~\cite{kalman1960}.

Kalman filtering has demonstrated robust real-time state estimation across diverse sensing applications, including adaptive multi-algorithm frameworks for structural gap detection using low-cost sensors~\cite{islam2025}.

Beyond the standard Kalman-filter update, ByteTrack's association step in this implementation extended the standard IoU-based association cost, \mbox{$C = 1 - \IoU$}, with a directional soft boost. The boost was developed specifically for this system's leftward-drift geometry rather than adapted from prior tracking literature, and was applied selectively to candidate pairs exhibiting leftward motion, consistent with the apparent leftward drift of stationary weeds through the camera's field of view as the platform advances. For track--detection pairs whose center displacement fell within $\pm30^{\circ}$ of horizontal in the leftward direction, a bonus term combining angular alignment and relative bounding-box area was added to the IoU similarity before rescaling:
\begin{align}
  b  &= 0.1\left(\frac{-0.5\,\lvert\theta\rvert}{30^{\circ}} + \kappa\right), \label{eq:bonus}\\
  s' &= \min\!\left(1,\ \frac{\IoU + b}{1.1}\right), \label{eq:simadj}
\end{align}
where $\theta$ is the displacement angle from horizontal and $\kappa$ is the ratio of the smaller to the larger bounding-box area of the candidate pair. The corresponding association cost is \mbox{$C = 1 - s'$}; pairs falling outside the leftward cone, including all rightward-moving candidates, retained the unmodified cost \mbox{$C = 1 - \IoU$}. This soft preferential weighting, rather than a hard exclusion of rightward matches, was intended to reduce association ambiguity between differently sized, closely spaced weeds under the platform's characteristic leftward apparent motion; no formal ablation quantifying its effect on track-switch or ID-switch rates was conducted in this study.

\subsubsection{Spray Timing and Actuation Control}
\label{sec:timing}
Tracked position and velocity estimates provide the inputs to the spray-timing model: each track's current position, combined with the current vehicle velocity estimate, determines when the corresponding nozzle must fire for the tracked plant to be centered beneath it at the moment of spray release. Precise spray timing accounts for physical camera-to-nozzle separation, vehicle velocity, and three fixed latency components: solenoid valve response ($\sim$18~ms, an assumed nominal value, not independently measured on this hardware), software processing (20.9~ms, measured; Table~\ref{tab:yolo}), and droplet flight time ($\sim$25~ms, also an assumed nominal value), together giving a total fixed delay of $\sim$64~ms; the assumed terms were not bench-validated on the study platform (see Section~\ref{sec:cropdamage}, Limitations). The spray trigger time delay $\Delta t$ for each nozzle is
\begin{equation}
\begin{aligned}
    \Delta t &= \frac{d}{v} + \tau_{\mathrm{L}},\\
    \tau_{\mathrm{L}} &= t_{\mathrm{valve}} + t_{\mathrm{sw}} + t_{\mathrm{flight}},
\end{aligned}
\label{eq:delay}
\end{equation}
where $d$ is the longitudinal camera-to-nozzle separation (8\nobreakdash--20~cm), $v$ is the current platform speed (0.13\nobreakdash--0.45~m/s), and $\tau_{\mathrm{L}} \approx 64$~ms is the total fixed latency: the assumed solenoid response time $t_{\mathrm{valve}}$ ($\sim$18~ms), the measured software-processing latency $t_{\mathrm{sw}}$ (20.9~ms; Table~\ref{tab:yolo}), and the assumed droplet travel time $t_{\mathrm{flight}}$ ($\sim$25~ms). This framework implies a speed-dependent positional error,
\begin{equation}
    \varepsilon = v\,\tau_{\mathrm{L}}.
    \label{eq:error}
\end{equation}
At 0.13~m/s (0.3~mph), the total latency of $\sim$64~ms produces $\varepsilon \approx 8.3$~mm. At 0.45~m/s (1.0~mph), the identical latency produces $\varepsilon \approx 29$~mm, a 3.5-fold increase consistent with the observed targeting degradation at higher speeds; note that the valve and droplet-flight terms are assumed nominal values rather than bench-measured quantities (see Section~\ref{sec:cropdamage}). The valve opening duration was calculated as
\begin{equation}
    t_{\mathrm{spray}} = 0.75\,\frac{w\,c_{\mathrm{px}}}{v},
    \label{eq:spray}
\end{equation}
where $w$ is the maximum bounding-box width (pixels), $c_{\mathrm{px}} = 0.838$~mm/pixel, and the factor of 0.75 prevents over-application; the valve duration was bounded between 50~ms and 200~ms.

Three post-processing filters were applied to raw detections before spray triggering. The detector's output classes reflect a multi-species label taxonomy retained for cross-season training compatibility; for field spray triggering, binary class merging remapped these predictions to two operational identities, weed (label ID 29) and crop (label ID 43, radish), since spray decisions depend only on this binary distinction rather than species identity. Region-of-interest filtering discarded detections falling outside the nozzle array's coverage width, since these could not be targeted regardless of detection confidence. Crop overlap filtering suppressed weed detections with IoU $> 0.25$ against a detected crop plant, preventing spray triggers for weeds growing in close proximity to crops where actuation risked crop contact.

\subsection{Experimental Validation}
\label{sec:expval}

\subsubsection{Indoor Validation}
\label{sec:indoor_val}
\begin{figure*}[!t]
    \centering
    \begin{tikzpicture}[inner sep=0pt]
      \node (img) {\includegraphics[trim=0 0 0 100bp,clip,width=0.80\textwidth]{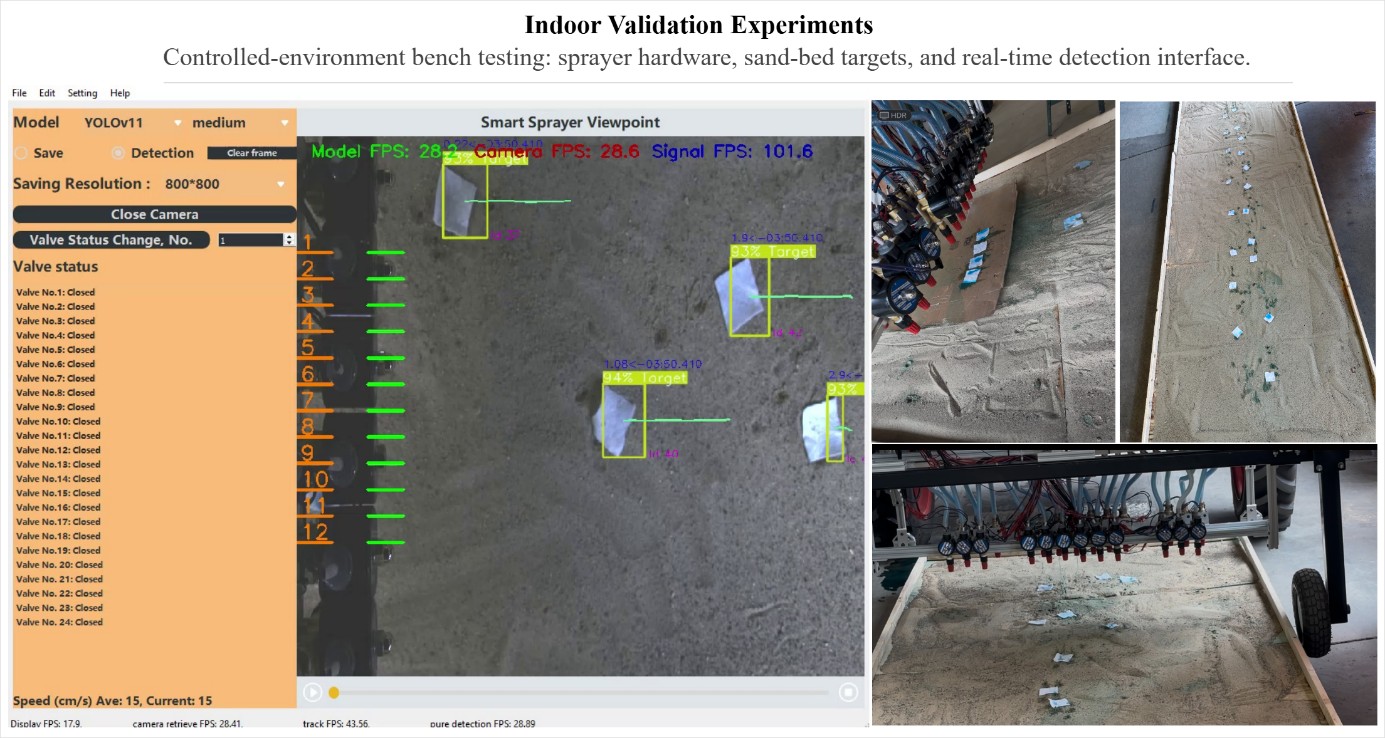}};
      \begin{scope}[x={($(img.south east)-(img.south west)$)},y={($(img.north west)-(img.south west)$)},shift={(img.south west)}]
        \node[ptag] at (0.006,0.992) {(a)};
        \node[ptag] at (0.634,0.992) {(b)};
      \end{scope}
    \end{tikzpicture}
    \caption{Indoor validation of the precision spraying system. (a)~Real-time software interface~\cite{deng2026a} showing the YOLOv11 detection output, the status of all 12 valves, and processing throughput (FPS). (b)~Sand bin test apparatus (0.6~m $\times$ 2.4~m) with 100 white polyester targets (5~cm $\times$ 5~cm) at approximately 5~targets/m$^2$; blue dye deposition after each pass was used to quantify weed hit rate and spray positioning accuracy.}
    \label{fig:indoor}
\end{figure*}

Initial system validation employed a sand bin test apparatus (0.6~m $\times$ 2.4~m) to establish baseline performance under controlled indoor conditions (Fig.~\ref{fig:indoor}), following a similar validation approach used in our group's prior sprayer prototypes~\cite{deng2026a,deng2026c}, adapted here for the reconfigured 12-nozzle array and updated detection model. One hundred white polyester targets (5~cm $\times$ 5~cm) were distributed at $\sim$5~targets/m$^2$, simulating field weed spatial arrangement. A simplified color-based detection algorithm (HSV thresholding with morphological operations) replaced YOLO to isolate spray hardware performance from detection variability. The vehicle traversed at 0.27~m/s (0.6~mph, mid-range of field speeds) using blue food-grade dye spray fluid ($\sim$5\% concentration) for visible deposition marking without phytotoxicity to test targets or, later, field crops. Post-traverse photography was conducted to document hit/miss status and spray positioning.

\subsubsection{Field Trials}
\label{sec:field_trials}
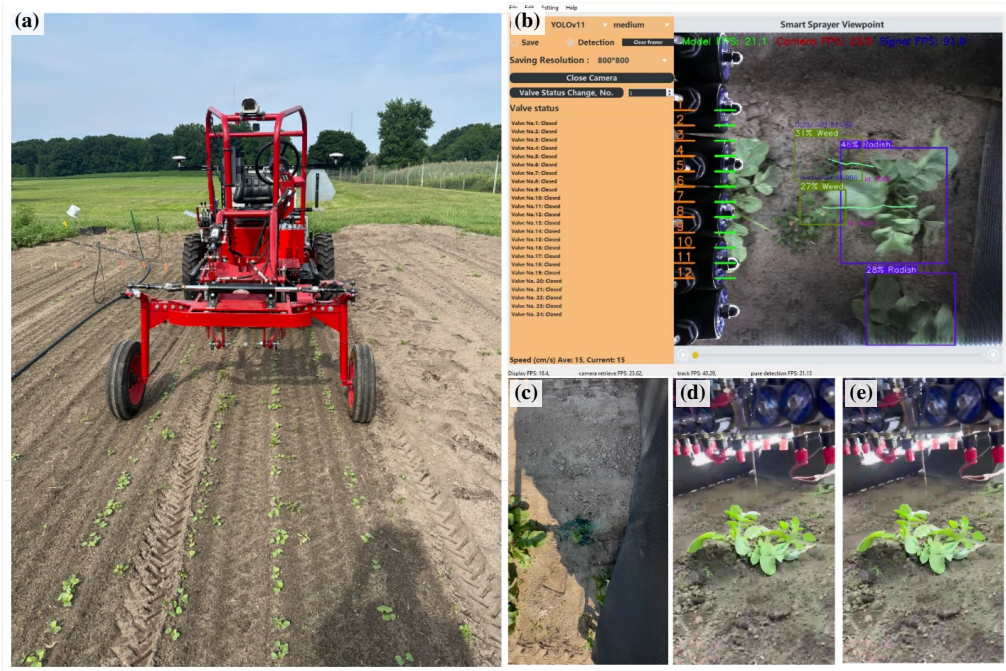
\begin{figure*}[!t]
    \centering
    \begin{tikzpicture}[inner sep=0pt]
      \node (img) {\includegraphics[trim=0 0 0 76bp,clip,width=0.74\textwidth]{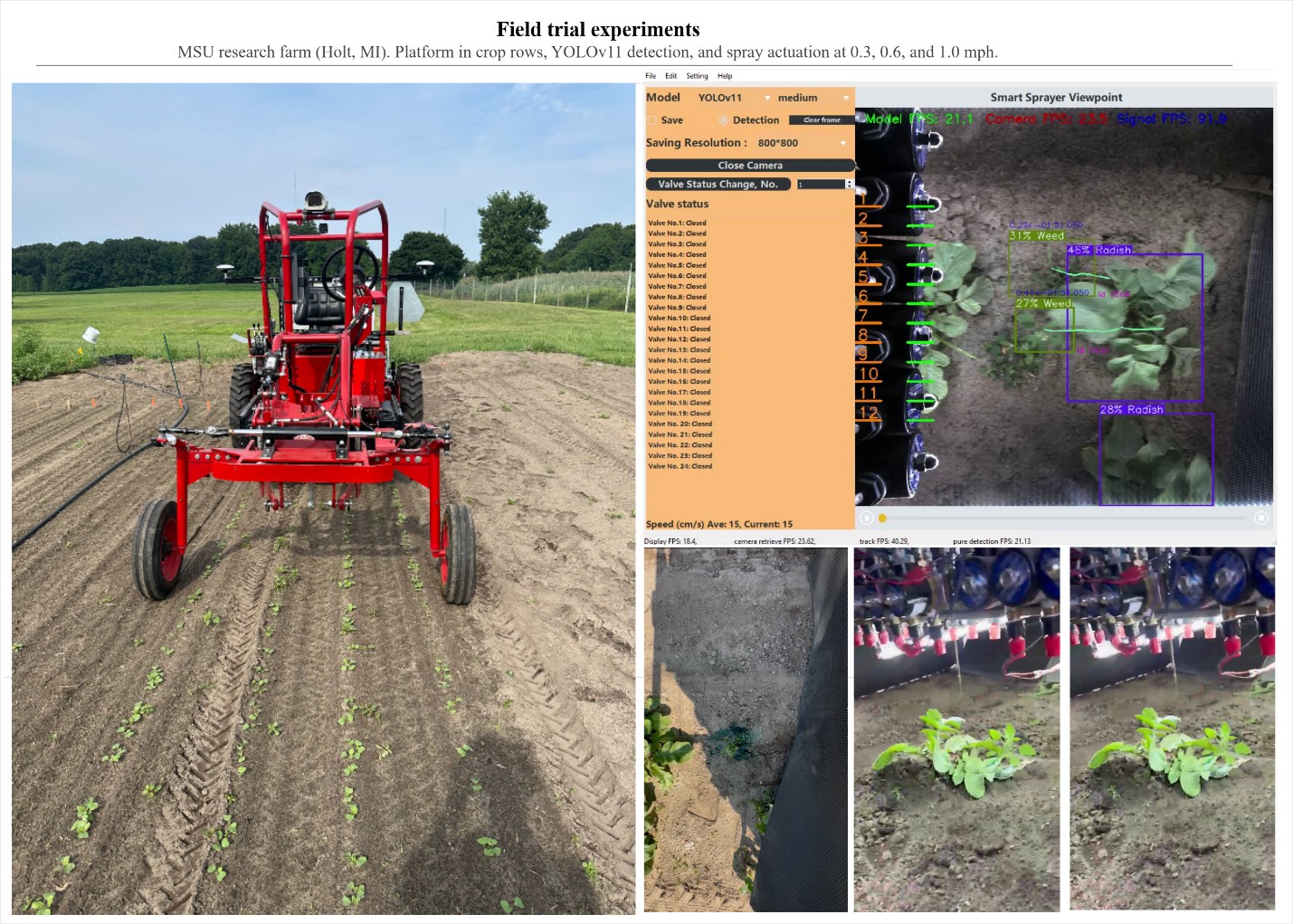}};
      \begin{scope}[x={($(img.south east)-(img.south west)$)},y={($(img.north west)-(img.south west)$)},shift={(img.south west)}]
        \node[ptag] at (0.008,0.992) {(a)};
        \node[ptag] at (0.500,0.992) {(b)};
        \node[ptag] at (0.500,0.438) {(c)};
        \node[ptag] at (0.663,0.438) {(d)};
        \node[ptag] at (0.830,0.438) {(e)};
      \end{scope}
    \end{tikzpicture}
    \caption{Field trials conducted in August 2025. (a)~The semi-autonomous spraying platform traversing radish rows at the MSU research farm. (b)~Real-time software interface showing YOLOv11 detections of weeds (green boxes) and radish (blue boxes). (c)--(e)~Close-up of inter-row weed populations targeted during the trials; blue dye deposition on soil and plant surfaces after a sprayer pass; and radish canopy at the 5--7 true-leaf stage.}
    \label{fig:field}
\end{figure*}

Field experiments were conducted at Michigan State University Horticulture Teaching and Research Center (42.7284$^{\circ}$N, 84.4812$^{\circ}$W, Holt, MI) on Capac loam soil (pH 6.3, 2.8\% organic matter), as shown in Fig.~\ref{fig:field}. Radish (\textit{Raphanus sativus} `Cherry Belle') was direct-seeded at 2.5~cm in-row $\times$ 38~cm between-row spacing in mid-July 2025. At spray trials (mid-August 2025, $\sim$4~weeks post-emergence), radish exhibited 5\nobreakdash--7 true leaves with 5\nobreakdash--10~cm canopy width, representing the final window for post-emergence herbicide application before canopy closure.

The experimental plot comprised two 30~m parallel rows with the system traversing the inter-row space. Spray coverage width ($\sim$13~cm) targeted zones immediately flanking crop rows where weed competition was most intense. Environmental conditions: 22\nobreakdash--28$^{\circ}$C air temperature, 55\nobreakdash--70\% relative humidity, $<$5~km/h wind, overcast to partly cloudy, moderate soil moisture ($\sim$25\% volumetric water content), no rainfall within 24~hours. Natural weed populations provided spray targets. Dominant species were common lambsquarters (\textit{Chenopodium album} L.) and common purslane (\textit{Portulaca oleracea} L.), both summer annuals characteristic of Michigan vegetable fields. Weed growth stages ranged from cotyledon to 5+ true leaves (5\nobreakdash--25~cm height, 2.5\nobreakdash--7.5~cm canopy width) at moderate density (5\nobreakdash--10~plants/m$^2$). Minimum detectable weed size, as constrained by the imaging and motion-blur characteristics described in Section~\ref{sec:vision}, was 2.5~cm canopy diameter.

The platform was operated under direct manual operator control for throttle throughout all trials, with auto-steering (Section~\ref{sec:platform}) maintaining row alignment. Field trials systematically evaluated performance across three speeds, tested in increasing order on the same pair of experimental rows (0.13, then 0.27, then 0.45~m/s; 0.3, 0.6, then 1.0~mph), characterizing degradation from slow research speeds toward the 0.67~m/s (1.5~mph) commercial threshold identified by \Vijay{} \textit{et al.}~\cite{vijayakumar2025}, whose reported optimal operating range for tomato and pepper systems (0.45\nobreakdash--0.89~m/s) informed the upper end of the tested range.

For each speed, the vehicle completed a single continuous pass with YOLOv11 detection and ByteTrack tracking active, using the same dilute blue food-grade dye spray fluid ($\sim$5\% concentration) described in Section~\ref{sec:indoor_val}, with approximately 3\nobreakdash--4~h separating consecutive speed treatments; residual dye from earlier passes was not visually distinguishable against soil and plant surfaces after this interval, so contamination of ground-truth scoring for later passes was not expected. Because all three speeds were tested sequentially on the same rows rather than on independent plots, the reported degradation trend reflects a single experimental unit assessed over time rather than replicated spatial units; this design and its implications are discussed as a limitation in Section~\ref{sec:cropdamage}.

\subsection{Performance Evaluation}
\label{sec:perfeval}

\subsubsection{Detection Performance Metrics}
\label{sec:detmetrics}
Detection performance was evaluated using three standard object-detection metrics, computed at $\IoU = 0.5$ and confidence $\geq 0.25$: precision, the fraction of detections correctly matching a ground-truth object; recall, the fraction of ground-truth objects successfully detected; and mAP@50, the mean average precision across $N$ classes at $\IoU = 0.5$ using 101-point COCO interpolation~\cite{lin2014}:
\begin{equation}
\begin{aligned}
    \mathrm{Precision} &= \frac{\mathrm{TP}}{\mathrm{TP}+\mathrm{FP}}\times 100\%,\\
    \mathrm{Recall}    &= \frac{\mathrm{TP}}{\mathrm{TP}+\mathrm{FN}}\times 100\%,\\
    \mAP_{50}          &= \frac{1}{N}\sum_{c=1}^{N}\mathrm{AP}_{c},
\end{aligned}
\label{eq:detmetrics}
\end{equation}
where TP, FP, and FN are the numbers of true-positive, false-positive, and false-negative detections, respectively, and $\mathrm{AP}_{c}$ is the average precision of class $c$ at $\IoU = 0.5$.

Dataset-based evaluation utilized the held-out test partition (20\% of annotated images) representing ideal imaging conditions. Video-based field evaluation extracted frames at $\sim$1~FPS (every 30th frame) from recorded streams, a practical sampling rate chosen to keep manual ground-truth annotation tractable across all three speed treatments rather than one derived from a formal detection-frequency analysis, yielding 70\nobreakdash--250 frames per speed treatment. Ground truth annotations for field frames were created using identical protocols as for training data (LabelImg tool, YOLO format). Video-based metrics provide conservative performance estimates, as tracking can recover missed detections through temporal association.

\subsubsection{Spray Targeting and Crop Protection Metrics}
\label{sec:spraymetrics}
Spray targeting accuracy was quantified through spatial analysis of blue dye deposition relative to ground truth plant locations, using the metrics summarized in Table~\ref{tab:metrics}. Spray area (SA), weed area (WA), and crop area (CA) were defined by bounding box extents, with centers computed as centroid coordinates. Separate calculations for crop and weed detection enabled discrimination between failure modes affecting targeting versus crop protection. Weed Hit Rate (WHR) serves as the operational proxy for weed control efficacy throughout this study; the commercial threshold of $>$90\% weed control efficacy (Table~\ref{tab:thresholds}) is equivalent to WHR $> 90\%$.

\begin{table*}[!t]
\centering
\caption{Spray Targeting and Crop Protection Metrics Used for System Evaluation}
\label{tab:metrics}
\footnotesize
\renewcommand{\arraystretch}{1.5}
\begin{tabularx}{\textwidth}{@{}>{\raggedright\arraybackslash}p{1.50in}>{\raggedright\arraybackslash}X>{\raggedright\arraybackslash}p{2.20in}@{}}
\toprule
Metric & Definition & Formula \\
\midrule
Weed hit rate (WHR) & Percentage of ground-truth weeds receiving spray contact ($\IoU > 0$). &
  $\displaystyle \WHR = \frac{N_{w}^{\mathrm{hit}}}{N_{w}^{\mathrm{gt}}}\times 100\%$ \\
Crop contact rate (CCR) & Percentage of ground-truth crops receiving any spray contact. &
  $\displaystyle \CCR = \frac{N_{c}^{\mathrm{hit}}}{N_{c}^{\mathrm{gt}}}\times 100\%$ \\
Weed precision rate (WPR) & Percentage of total spray area deposited on weed targets. &
  $\displaystyle \mathrm{WPR} = \frac{A_{w}}{A_{T}}\times 100\%$ \\
Crop precision rate (CPR) & Percentage of total spray area deposited on crop plants. &
  $\displaystyle \mathrm{CPR} = \frac{A_{c}}{A_{T}}\times 100\%$ \\
Soil precision rate (SPR) & Percentage of total spray area deposited on bare soil. &
  $\displaystyle \mathrm{SPR} = \frac{A_{b}}{A_{T}}\times 100\%$ \\
Average aligned distance ($d_{\mathrm{avg}}$) & Mean Euclidean distance between the spray centroid and the weed centroid for successfully hit weeds (mm). &
  $\displaystyle d_{\mathrm{avg}} = \frac{1}{N}\sum_{i=1}^{N}\bigl\lVert \mathrm{SC}_{i}-\mathrm{WC}_{i}\bigr\rVert$ \\
Average IoU ($\IoU_{\mathrm{avg}}$) & Mean intersection over union between the spray area and the weed area for successfully hit weeds. &
  $\displaystyle \IoU_{\mathrm{avg}} = \frac{1}{N}\sum_{i=1}^{N}\frac{\lvert\mathrm{SA}_{i}\cap\mathrm{WA}_{i}\rvert}{\lvert\mathrm{SA}_{i}\cup\mathrm{WA}_{i}\rvert}$ \\
Translation efficiency ($\eta_{\mathrm{trans}}$) & Ratio of weed hit rate to detection recall; quantifies the cumulative loss through the detection--tracking--actuation pipeline. &
  $\displaystyle \eta_{\mathrm{trans}}(v) = \frac{\WHR(v)}{\mathrm{Recall}(v)}\times 100\%$ \\
Selectivity ratio ($S$) & Ratio of weed hit rate to crop contact rate; measures discrimination between target and protected plants. &
  $\displaystyle S(v) = \frac{\WHR(v)}{\CCR(v)}$ \\
\bottomrule
\end{tabularx}
\tabnote{\textit{Notation:}
$N_{w}^{\mathrm{gt}}$, $N_{c}^{\mathrm{gt}}$: total numbers of ground-truth weed and crop instances;
$N_{w}^{\mathrm{hit}}$, $N_{c}^{\mathrm{hit}}$: numbers of weed and crop instances receiving spray contact;
$A_{T} = A_{w} + A_{c} + A_{b}$: total spray area, decomposed into the areas deposited on weeds, crops, and bare soil;
$N$: number of successfully hit weeds used in the distance and IoU averages;
$\mathrm{SC}_{i}$, $\mathrm{WC}_{i}$: spray and weed centroids of the $i$-th successfully hit weed;
$\mathrm{SA}_{i}$, $\mathrm{WA}_{i}$: spray and weed areas of the $i$-th successfully hit weed;
$v$, $\mathrm{Recall}(v)$: travel speed and the corresponding object-detection recall at that speed.
Precision rates (WPR, CPR, SPR) may sum to more than 100\% because a single spray event can contact several surface types simultaneously.}
\end{table*}

Commercial viability thresholds were based on agronomic requirements and operational economics (Table~\ref{tab:thresholds}). These represent concurrent requirements; all three criteria must be satisfied simultaneously, as a deficiency in any dimension renders the system unsuitable for production agriculture.

\begin{table*}[!t]
\centering
\caption{Commercial Viability Thresholds for Precision Spraying Systems}
\label{tab:thresholds}
\footnotesize
\renewcommand{\arraystretch}{1.2}
\begin{tabularx}{\textwidth}{@{}>{\raggedright\arraybackslash}p{1.35in}>{\centering\arraybackslash}p{1.15in}>{\raggedright\arraybackslash}X>{\raggedright\arraybackslash}p{1.75in}@{}}
\toprule
Criterion & Threshold & Rationale & Source \\
\midrule
Weed control efficacy & $>$90\% & Conventional herbicides achieve 90\nobreakdash--99\% control; precision systems must approach this standard to minimize supplemental manual weeding. & Slaughter \textit{et al.}~\cite{slaughter2008} \\
Crop damage & $<$10\% & Economically acceptable yield-loss threshold for vegetable production with narrow profit margins. & Slaughter \textit{et al.}~\cite{slaughter2008} \\
Operating speed & $\geq 0.67$~m/s (1.5~mph) & Minimum for commercial field efficiency in specialty-crop operations; enables timely weed control within critical windows. & \Vijay{} \textit{et al.}~\cite{vijayakumar2025} \\
\bottomrule
\end{tabularx}
\end{table*}

\subsubsection{Analytical Approach and Ground-Truth Methodology}
\label{sec:analytical}
Performance analyses addressed three objectives: indoor--field performance gap quantification, speed-dependent degradation characterization, and failure mode categorization. All analyses employed descriptive statistics (means, standard deviations, ranges) as the experimental design focused on comprehensive characterization of a single system across controlled speed treatments rather than hypothesis testing. Given that each speed treatment comprised a single continuous pass on the same experimental rows rather than replicated, independent trials (Section~\ref{sec:field_trials}), inferential statistics were not applicable; characterization was the primary analytical objective.

Indoor--field comparison quantified the performance penalty of transitioning from controlled laboratory to field conditions by computing ratios and absolute differences in detection accuracy, weed hit rate, and targeting metrics at matched speed (0.27~m/s, 0.6~mph).

Speed-dependent analysis characterized degradation trajectories by plotting each metric against operating speed (0.13, 0.27, 0.45~m/s; 0.3, 0.6, 1.0~mph). Linear regression quantified degradation rates (performance change per m/s). The detection-to-spraying performance gap was calculated at each speed to quantify pipeline losses through the detection--tracking--actuation sequence. Given that only three speed treatments were evaluated without replication, regression models represent descriptive trend estimates rather than inferential statistical results.

Failure mode categorization classified unsuccessful spray events (missed weeds, crop contacts) into diagnostic categories: detection failure (weed undetected), tracking failure (track lost before triggering), timing failure (spray displaced $>$20~mm longitudinally), targeting failure (lateral offset despite correct timing), and intentional crop-proximity suppression (IoU $> 0.25$ with crops). Failure mode proportions were calculated separately per speed to characterize shifting bottlenecks. Complete failure mode taxonomy appears in Supplementary Table S3.

Ground truth for spray performance was established through paired pre- and post-spray photography. The forward-facing detection camera described in Section~\ref{sec:vision} supplied the pre-spray reference frame for each plant, captured as it entered the field of view ahead of the nozzle array. A second RGB camera, mounted at the rear of the platform immediately behind the spray extension, operated continuously at 30~FPS and captured the corresponding post-spray frame as each plant emerged from beneath the nozzle array.

Pre- and post-spray frames were matched by ground location using row-relative distance markers and imported into ImageJ for spray footprint digitization: for each matched pair, the spray deposit was traced to compute its area and centroid, and this centroid was compared against the corresponding plant centroid from the pre-spray frame to determine whether the deposit fell on the intended weed, an adjacent crop plant, or bare soil. Pixel measurements were converted to millimeters using the calibrated ground-sampling distance, yielding the aligned distance and IoU metrics reported in Table~\ref{tab:metrics}. The resulting hit/miss classification for every detected weed and crop plant in the field trials formed the ground-truth basis for the weed hit rate (WHR) and crop contact rate (CCR) calculations reported in Section~\ref{sec:fieldspray}.

To estimate performance at the untested commercial threshold speed, linear regression was fit to each metric (mAP@50, WHR, CCR) as a function of operating speed across the three tested conditions (Section~\ref{sec:benchmarks}).

\section{Results}
\label{sec:results}

\subsection{Overview}
\label{sec:overview}
Indoor validation at 0.27~m/s (0.6~mph) using simulated targets established baseline capabilities under controlled conditions. Field trials at three operating speeds (0.13, 0.27, and 0.45~m/s; 0.3, 0.6, and 1.0~mph) characterized speed-dependent degradation of detection accuracy, targeting precision, and crop selectivity in radish production with natural weed populations, enabling systematic comparison against commercial viability requirements. Sections~\ref{sec:dataperf}--\ref{sec:indoor_perf} report dataset-based and indoor baseline detection performance, addressing objective~(i). Sections~\ref{sec:fielddet}--\ref{sec:fieldspray} characterize speed-dependent field degradation across the detection--tracking--actuation pipeline, including crop damage, addressing objective~(ii). Section~\ref{sec:benchmarks} compares field performance against commercial viability thresholds and extrapolates degradation trends to the untested commercial speed, addressing objective~(iii). Implications for multi-season generalization and the improvement roadmap are taken up in Section~\ref{sec:discussion} (Discussion).

\subsection{Dataset-Based Detection Performance}
\label{sec:dataperf}
Three YOLOv11 variants were evaluated on the held-out test partition of the five-season dataset (Table~\ref{tab:yolo}). This benchmark was first reported by Deng \textit{et al.}~\cite{deng2026c}; it is reproduced here for completeness and to contextualize the field-deployment results that follow. YOLOv11-medium achieved the highest accuracy (89.9\% mAP@50, 20.9~ms inference) and was selected for field deployment based on superior performance while maintaining real-time processing compatibility at 30~FPS. Overall mAP@50 for nano, small, and medium was 86.2\%, 87.2\%, and 89.9\%, respectively. Beyond the mAP@50 values summarized in Table~\ref{tab:yolo}, test-set analysis of the deployed medium model revealed distinct error patterns: weed precision (91.0\%) exceeded recall (79.3\%), while crop detection showed more balanced performance (88.9\% precision, 88.4\% recall).

\begin{table}[!t]
\centering
\caption{YOLOv11 Variant Performance on the Five-Season Dataset}
\label{tab:yolo}
\footnotesize
\setlength{\tabcolsep}{5pt}
\renewcommand{\arraystretch}{1.12}
\begin{tabular}{@{}lccc@{}}
\toprule
Model & \makecell{Inference\\time (ms)} & \makecell{Crops\\mAP@50 (\%)} & \makecell{Weeds\\mAP@50 (\%)} \\
\midrule
YOLOv11-nano & 10.0 & 89.8 & 82.5 \\
YOLOv11-small & 14.6 & 87.1 & 87.2 \\
YOLOv11-medium & 20.9 & 90.7 & 89.1 \\
\bottomrule
\end{tabular}
\tabnote{Evaluated on the held-out test partition (20\% of the combined five-season dataset). mAP@50 = mean average precision at an IoU threshold of 0.5. Inference time was measured on an NVIDIA RTX 4060 Ti GPU.}
\end{table}

\subsection{Indoor Validation Performance}
\label{sec:indoor_perf}
Indoor validation at 0.27~m/s (0.6~mph) demonstrated near-optimal performance (Tables~\ref{tab:detspeed} and~\ref{tab:sprayspeed}). Detection using the color-based baseline algorithm achieved 99.1\% AP@50 (bounding-box level, simulated white targets). Note that indoor AP@50 was computed on simulated white targets using color thresholding; it is not directly comparable to YOLOv11 field mAP@50 and serves only to isolate spray hardware performance from detection variability.

Spraying achieved an 82.0\% target hit rate with a 7.5~mm mean aligned distance and IoU of 0.53 (Table~\ref{tab:sprayspeed}). The 18\% miss rate despite 99.1\% AP@50 demonstrated that spray timing limitations, not detection failures, constrained performance. Missed targets were distributed across nozzle positions without systematic spatial patterns.

\subsection{Field Detection Performance}
\label{sec:fielddet}
Field detection performance degraded progressively with speed (Table~\ref{tab:detspeed}; Fig.~\ref{fig:speedacc}). Across 0.13\nobreakdash--0.45~m/s (0.3\nobreakdash--1.0~mph), mAP@50 declined 24.5 percentage points (94.6\% $\rightarrow$ 70.1\%), a 26\% relative decrease, with degradation accelerating at higher speeds: the 0.13\nobreakdash--0.27~m/s interval produced a 9.7-point decline versus 14.8 points for 0.27\nobreakdash--0.45~m/s, confirming non-linear performance loss. Recall degraded more severely than precision (30.7 vs.\ 19.5 percentage points), with 37.5\% of weeds undetected at 0.45~m/s compared to only 6.8\% at 0.13~m/s, indicating that weeds became undetectable under motion rather than misclassified. Indoor--field comparison at 0.27~m/s showed a 14.2-point performance reduction (99.1\% $\rightarrow$ 84.9\% mAP@50), while the relationship to dataset baseline shifted with speed: at 0.13~m/s, field mAP@50 (94.6\%) exceeded the static dataset performance (89.9\%), but at 0.45~m/s fell 19.8 points below it (70.1\% vs.\ 89.9\%), demonstrating that dataset-based metrics substantially overestimate field capability at higher operating speeds.

\begin{table}[!t]
\centering
\caption{Detection Performance Across Operating Speeds}
\label{tab:detspeed}
\footnotesize
\setlength{\tabcolsep}{5pt}
\renewcommand{\arraystretch}{1.12}
\begin{tabular}{@{}lccc@{}}
\toprule
Test condition & Precision (\%) & Recall (\%) & mAP@50 (\%) \\
\midrule
Indoor (0.27~m/s) & 98.1 & 99.5 & 99.1 \\
Field (0.13~m/s) & 95.8 & 93.2 & 94.6 \\
Field (0.27~m/s) & 88.4 & 80.7 & 84.9 \\
Field (0.45~m/s) & 76.3 & 62.5 & 70.1 \\
\bottomrule
\end{tabular}
\tabnote{Indoor validation used color-based detection on white targets; field validation used YOLOv11-medium on natural weed and crop populations. mAP@50 = mean average precision at an IoU threshold of 0.5. Speed equivalents: 0.13~m/s = 0.3~mph, 0.27~m/s = 0.6~mph, 0.45~m/s = 1.0~mph.}
\end{table}

\subsection{Field Spraying Performance}
\label{sec:fieldspray}
Weed hit rate declined progressively across the field speed range, from 78.3\% at 0.13~m/s (0.3~mph) to 35.0\% at 0.45~m/s (1.0~mph), a 55.3\% relative decline within the tested field range, and a 57.3\% relative decrease from the 82.0\% indoor baseline (Table~\ref{tab:sprayspeed}). Decline accelerated at higher speeds: a 19.6-point drop between 0.13\nobreakdash--0.27~m/s versus 23.7 points between 0.27\nobreakdash--0.45~m/s.

\begin{figure}[!t]
    \centering
    \begin{tikzpicture}
    \begin{axis}[
        width=\columnwidth, height=0.74\columnwidth,
        xmin=0.05, xmax=0.56, ymin=0, ymax=105,
        xtick={0.13,0.27,0.45},
        xticklabels={{0.13\\(0.3~mph)},{0.27\\(0.6~mph)},{0.45\\(1.0~mph)}},
        xticklabel style={align=center},
        ytick={0,20,40,60,80,100},
        yticklabel={\pgfmathprintnumber{\tick}\%},
        xlabel={Travel speed (m/s)}, ylabel={Performance (\%)},
        grid=major, grid style={gray!25,thin},
        tick label style={font=\scriptsize}, label style={font=\footnotesize},
        legend style={font=\scriptsize,at={(0.5,-0.36)},anchor=north,legend columns=-1,
                      /tikz/every even column/.append style={column sep=6pt}},
        clip=false,
    ]
    \addplot[draw=none,fill=ColG!14,forget plot] coordinates {(0.05,90)(0.56,90)(0.56,100)(0.05,100)} \closedcycle;
    \addplot[ColG,dashed,thick,no marks,forget plot] coordinates {(0.05,90)(0.56,90)};
    \addplot[ColC,dashed,thick,no marks,forget plot] coordinates {(0.05,10)(0.56,10)};
    \node[anchor=north east,font=\scriptsize,text=ColG] at (axis cs:0.555,99) {WHR target ($>$90\%)};
    \node[anchor=south east,font=\scriptsize,text=ColC] at (axis cs:0.555,10.8) {CCR limit ($<$10\%)};
    \addplot[ColW,thick,mark=*,mark size=1.8pt,nodes near coords,point meta=y,
        nodes near coords={\pgfmathprintnumber[fixed,zerofill,precision=1]{\pgfplotspointmeta}\%},
        every node near coord/.style={font=\scriptsize,anchor=east,xshift=-3pt,text=ColW}]
        coordinates {(0.13,78.3)(0.27,58.7)(0.45,35.0)};
    \addplot[ColC,thick,mark=square*,mark size=1.7pt,nodes near coords,point meta=y,
        nodes near coords={\pgfmathprintnumber[fixed,zerofill,precision=1]{\pgfplotspointmeta}\%},
        every node near coord/.style={font=\scriptsize,anchor=north,yshift=-2pt,text=ColC}]
        coordinates {(0.13,20.1)(0.27,24.3)(0.45,28.7)};
    \addplot[ColM,thick,dashed,mark=triangle*,mark size=2.0pt,nodes near coords,point meta=y,
        nodes near coords={\pgfmathprintnumber[fixed,zerofill,precision=1]{\pgfplotspointmeta}\%},
        every node near coord/.style={font=\scriptsize,anchor=south,yshift=1pt,text=ColM,fill=white,fill opacity=0.75,text opacity=1,inner sep=0.8pt}]
        coordinates {(0.13,94.6)(0.27,84.9)(0.45,70.1)};
    \legend{WHR,CCR,mAP@50}
    \draw[<->,>={Stealth[length=3pt,width=2.4pt]},ColG,thin] (axis cs:0.16,78.3) -- (axis cs:0.16,90)
      node[midway,right,font=\scriptsize,text=ColG,inner sep=1.5pt] {$-11.7$};
    \draw[<->,>={Stealth[length=3pt,width=2.4pt]},ColG,thin] (axis cs:0.30,58.7) -- (axis cs:0.30,90)
      node[midway,right,font=\scriptsize,text=ColG,inner sep=1.5pt] {$-31.3$};
    \draw[<->,>={Stealth[length=3pt,width=2.4pt]},ColG,thin] (axis cs:0.48,35.0) -- (axis cs:0.48,90)
      node[midway,right,font=\scriptsize,text=ColG,inner sep=1.5pt] {$-55.0$};
    \end{axis}
    \end{tikzpicture}
    \caption{Speed--accuracy trade-offs across the three operating speeds in radish production: weed hit rate (WHR), crop contact rate (CCR), and detection accuracy (mAP@50). Dashed lines mark the commercial-viability thresholds (WHR $>$90\%, CCR $<$10\%); vertical arrows give the shortfall of WHR relative to the 90\% target in percentage points.}
    \label{fig:speedacc}
\end{figure}
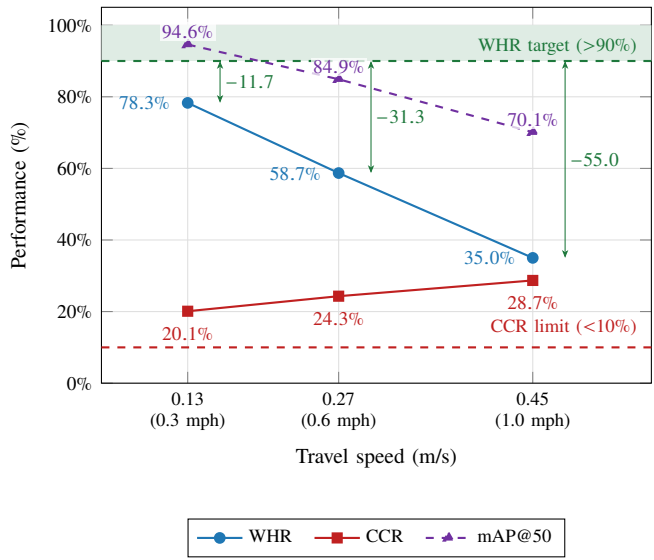

Crop contact rates increased with speed (20.1\% $\rightarrow$ 28.7\%), exceeding the 10\% commercial tolerance at all speeds. Weed:crop selectivity ratios (Table~\ref{tab:sprayspeed}) declined from 3.9:1 to 1.2:1. Spray deposition increasingly contacted multiple surfaces: at 0.45~m/s (1.0~mph), precision rates summed to 152.6\% (71.6\% weeds, 38.2\% crops, 42.8\% soil). Fig.~\ref{fig:speedacc} summarizes these speed--accuracy trade-offs, plotting weed hit rate, crop contact rate, and detection accuracy together against travel speed relative to the commercial viability thresholds established in Section~\ref{sec:perfeval}. Representative footage of live spray actuation and field-based weed detection and tracking during the multi-speed trials is provided in Supplementary Video S1~\cite{islam2026video}.

Spatial targeting degraded substantially with platform speed. Weed-aligned distance increased 2.1-fold (10.2 $\rightarrow$ 21.3~mm), while weed IoU declined 50\% (0.16 $\rightarrow$ 0.08). Crop targeting showed similar but less severe degradation (1.5-fold distance increase, 36\% IoU decline). The detection-to-spraying performance gap (mAP@50 minus weed hit rate) widened with speed: 16.3 points at 0.13~m/s (0.3~mph), 26.2 points at 0.27~m/s (0.6~mph), and 35.1 points at 0.45~m/s (1.0~mph). Translation efficiency (as defined in Table~\ref{tab:metrics}) declined from 84\% to 56\% across the tested speed range (Table~\ref{tab:sprayspeed}).

\begin{table}[!t]
\centering
\caption{Spraying Performance Across Operating Speeds}
\label{tab:sprayspeed}
\footnotesize
\setlength{\tabcolsep}{4pt}
\renewcommand{\arraystretch}{1.1}
\begin{tabular}{@{}lcccc@{}}
\toprule
 & Indoor & \multicolumn{3}{c}{Field} \\
\cmidrule(lr){2-2}\cmidrule(l){3-5}
Metric & 0.27~m/s & 0.13~m/s & 0.27~m/s & 0.45~m/s \\
\midrule
Weed hit rate (\%) & 82.0 & 78.3 & 58.7 & 35.0 \\
Crop contact rate (\%) & --- & 20.1 & 24.3 & 28.7 \\
Translation efficiency (\%) & 82.4 & 84.0 & 72.7 & 56.0 \\
Selectivity ratio & --- & 3.9:1 & 2.4:1 & 1.2:1 \\
\midrule
Weed precision rate (\%) & 77.0 & 90.1 & 82.4 & 71.6 \\
Crop precision rate (\%) & --- & 28.5 & 33.1 & 38.2 \\
Soil precision rate (\%) & 23.0 & 22.3 & 31.5 & 42.8 \\
\midrule
Weed-aligned distance (mm) & 7.5 & 10.2 & 15.7 & 21.3 \\
Crop-aligned distance (mm) & --- & 36.1 & 31.8 & 38.5 \\
Spray-to-weed IoU & 0.53 & 0.16 & 0.12 & 0.08 \\
Spray-to-crop IoU & --- & 0.11 & 0.09 & 0.07 \\
\bottomrule
\end{tabular}
\tabnote{Translation efficiency and selectivity ratio are defined in Table~\ref{tab:metrics}. Indoor validation used simulated white targets; crop metrics are not applicable (---). Precision rates may sum to more than 100\% because spray events can contact several surface categories. Aligned distance = Euclidean distance between spray center and target center; IoU = intersection over union between spray footprint and target area. Speed equivalents: 0.13~m/s = 0.3~mph, 0.27~m/s = 0.6~mph, 0.45~m/s = 1.0~mph.}
\end{table}

\subsection{Comparison with Commercial Viability Benchmarks}
\label{sec:benchmarks}
Commercial viability requires concurrent achievement of three criteria established in the precision spraying literature: weed control efficacy exceeding 90\%, matching the standard achieved by conventional broadcast herbicides~\cite{slaughter2008}; crop damage below 10\%, representing the economically acceptable yield loss threshold for vegetable production with narrow profit margins~\cite{slaughter2008}; and operating speed of at least 0.67~m/s (1.5~mph), the minimum for field efficiency in specialty crop operations identified by \Vijay{} \textit{et al.}~\cite{vijayakumar2025}. All three criteria must be satisfied simultaneously, as a deficiency in any single dimension renders the system unsuitable for commercial production.

The best field performance achieved in this study, at 0.13~m/s (0.3~mph), nonetheless did not meet the three criteria. Weed hit rate reached only 78.3\%, falling 11.7 percentage points short of the 90\% target. Crop contact rate was 20.1\%, exceeding the 10\% commercial tolerance by 10.1 percentage points. Operating speed itself fell short by the widest margin: even at its best-performing (and slowest) tested speed, 0.13~m/s, the system operated 80.6\% slower than the 0.67~m/s (1.5~mph) minimum required for commercial field efficiency.

These gaps are best read diagnostically rather than as a verdict on the approach itself. Detection accuracy remained comparatively stable across the tested range (mAP@50: 94.6\% at 0.13~m/s, declining by roughly 24 points to 0.45~m/s), while weed hit rate collapsed more than 43 percentage points over the same interval. That asymmetry is the central finding of this evaluation: the shortfall against commercial thresholds originates downstream of detection, in the tracking-to-actuation stages that convert a detection into a spray event, not in perception itself. Locating the bottleneck there, rather than in the vision model, points to where the next iteration should focus.

Velocity estimation uncertainty of 5\nobreakdash--10\% from wheel-slip-based speed sensing amplifies spray-timing position error 3.1-fold across the tested range (Section~\ref{sec:pipeline}), a mechanism unrelated to detection accuracy and correctable through RTK-GPS-based velocity estimation and faster valve actuation. Closing the gap to commercial thresholds is hence an actuation- and timing-engineering problem at this point, rather than a detection problem, a distinction that single-speed, detection-only evaluations elsewhere in the literature cannot make.

Linear regression across the three field speeds quantifies degradation rates and enables extrapolation to the commercial threshold. Note that $R^2$ values approaching 1.0 are mathematically expected when fitting a line to only three data points (Fig.~\ref{fig:speedacc}) and should not be interpreted as strong statistical validation of linearity; these models serve as simple first-order approximations only:
\begin{align}
  \widehat{\mAP}_{50}(v) &= 105.0 - 76.8\,v, & R^{2} &= 0.998, \label{eq:reg_map}\\
  \widehat{\WHR}(v)      &= 95.6 - 135.1\,v,  & R^{2} &= 1.000, \label{eq:reg_whr}\\
  \widehat{\CCR}(v)      &= 16.8 + 26.8\,v,   & R^{2} &= 0.997, \label{eq:reg_ccr}
\end{align}
where $v$ is platform velocity in m/s and outputs are in percentage points. Extrapolating to 0.67~m/s (1.5~mph) predicts a further severe collapse in performance, with weed hit rate falling to $\approx$5\% (Table~\ref{tab:extrap}), rendering the system agriculturally ineffective at commercial operating speed.

\begin{table}[!t]
\centering
\caption{Extrapolated Performance at 0.67~m/s}
\label{tab:extrap}
\footnotesize
\setlength{\tabcolsep}{4pt}
\renewcommand{\arraystretch}{1.12}
\begin{tabularx}{\columnwidth}{@{}lccc@{}}
\toprule
Metric & \makecell{Best tested\\(0.13~m/s)} & \makecell{Extrapolated\\(0.67~m/s)} & \makecell{Commercial\\threshold} \\
\midrule
mAP@50 (\%) & 94.6 & 53.5 & --- \\
WHR (\%) & 78.3 & 5.1 & $>$90 \\
CCR (\%) & 20.1 & 34.7 & $<$10 \\
\bottomrule
\end{tabularx}
\tabnote{Extrapolated values assume that linear degradation continues beyond the tested range; actual performance may degrade faster because of non-linear motion effects. WHR = weed hit rate; CCR = crop contact rate.}
\end{table}

Across the tested speed range, performance degraded through four compounding patterns. Detection recall declined steadily, falling 30.7 percentage points from the slowest to the fastest speed. Crop contact rates remained elevated at 20\nobreakdash--29\% across all three speeds, consistently exceeding the 10\% commercial tolerance even under the most favorable conditions. The gap between detection and spraying performance widened substantially with speed, growing from 16.3 to 35.1 percentage points as translation-efficiency losses accumulated through the tracking--actuation pipeline. Spatial targeting also deteriorated: aligned distance increased 2.1-fold and spray-to-weed IoU fell by half, indicating that even successfully triggered sprays landed less precisely on their intended targets as speed increased.

\section{Discussion}
\label{sec:discussion}

\subsection{Compounding Pipeline Losses Drive Speed-Dependent Degradation}
\label{sec:pipeline}
The central finding of this multi-speed field characterization challenges a widely held assumption: that improving detection accuracy is the primary path to commercial viability. Across the tested speed range, detection recall declined 30.7 percentage points while weed hit rate declined 43.3 percentage points, and translation efficiency (the fraction of detected weeds actually sprayed) fell from 84\% to 56\% (Table~\ref{tab:sprayspeed}), confirming that errors compound multiplicatively through the detection--tracking--actuation pipeline rather than residing in any single stage. Four primary limitations drove this: detection recall degradation under motion, excessive crop contact at all speeds, widening spray timing errors with speed, and degraded spatial targeting.

These findings also challenge the broader literature's reliance on single-speed evaluations, which systematically obscures speed sensitivity and creates an overly optimistic picture of operational readiness; the systematic three-speed design used here instead isolates where in the pipeline performance actually breaks down, providing the evidentiary basis for the prioritized improvement roadmap developed below rather than a generic call for ``better detection''.

\Vijay{} \textit{et al.}~\cite{vijayakumar2025} determined their reported 0.45\nobreakdash--0.89~m/s operating range through a systematic frame-capture analysis across three tested speeds (0.45, 0.89, and 1.34~m/s, each repeated twice), establishing via RTK-GPS-synced frame counts that at least six to seven continuous frames containing the target weed were needed for reliable actuation, a criterion satisfied up to 0.89~m/s but not above 1.12~m/s. Notably, their lowest tested speed (0.45~m/s) corresponds to this study's highest tested speed, at which our own detection and spraying performance had already degraded substantially; the two systems' validated operating envelopes therefore do not overlap under the conditions tested here.

This gap is not evidence that speed tolerance is an arbitrary, non-generalizable property of each system, but reflects a concrete, addressable difference in engineering approach: \Vijay{} \textit{et al.}'s system combined proportional-valve control with a frame-count-based targeting criterion specifically tuned to sustain reliable actuation at higher speeds, whereas the present system's fixed, non-speed-adaptive latency budget (Section~\ref{sec:timing}) and declining frame density (Eq.~(\ref{eq:density})) were not engineered for sustained operation above roughly 0.3~m/s. Closing this gap is therefore an engineering target for the roadmap in Section~\ref{sec:cropdamage}, not evidence that robust, speed-tolerant precision spraying is unattainable.

The speed-dependent degradation reflects compounding physical constraints that persist regardless of algorithm choice. Frame density dropped 71\% across the tested range, reducing per-plant detection opportunities from 230 to 67 frames per meter, while motion blur increased pixel displacement 3.4-fold, rendering sub-centimeter morphological features of small weeds indistinguishable from background texture. The differential recall-versus-precision pattern confirms weeds became undetectable rather than misclassified, with precision remaining above 76\% even at the fastest speed.

These detection losses then propagate and amplify through the actuation pipeline: fixed system latencies totaling $\sim$64~ms produce position errors that scale linearly with speed ($\approx$8~mm at 0.13~m/s to $\approx$29~mm at 0.45~m/s, a 3.5-fold increase; Section~\ref{sec:timing}), shifting spray from target center to edge or beyond. ByteTrack's two-stage association proved effective for handling partial occlusion and confidence fluctuations, but tracking gains were limited because spray timing constraints dominated; weed hit rates consistently fell below corresponding recall values. Together, these mechanisms explain why even a perfected detection model would still face substantial spraying degradation at commercial speeds. The improvement roadmap must prioritize hardware interventions, such as higher-frame imaging, the use of fast-responsive pulse-width modulation (PWM) valves in place of the present valve ($\sim$18~ms solenoid response), and more robust platform speed estimates independent of wheel slip.

Field-based detection performance at the slowest platform speed exceeded the dataset-based baseline (94.6\% vs.\ 89.9\% mAP@50), likely reflecting ByteTrack's temporal integration recovering missed single-frame detections. This is also consistent with the diversity of the five-season dataset used for model training. However, this advantage diminished at higher platform speeds, where motion-induced image degradation (e.g., blur), absent from static training images, remains a distinct and unresolved challenge.

A comparable dataset-to-field performance gap has not been explicitly quantified in our prior work~\cite{deng2026a,deng2026c}. Here, the 19.8~percentage-point gap between dataset-based detection (89.9\% mAP@50; Table~\ref{tab:yolo}) and video-based field performance at the highest platform speed (70.1\%; Table~\ref{tab:detspeed}) highlights the limitations of static image benchmarks in predicting performance degradation under continuous field motion.

This finding is further supported by Deng \textit{et al.}~\cite{deng2026b}, who evaluated a higher-precision micro-jet sprayer in a lettuce field at a similar platform speed ($\sim$0.25~m/s). Although the system achieved a field detection accuracy of 80.0\% mAP@50, its crop hit rate of 17.6\% still exceeded the 10\% commercial tolerance despite finer actuation. This is further consistent, using different hardware, with the need for field-based validation to anticipate real-world degradation, and with crop-adjacent weed targeting remaining an open problem requiring continued investigation.

\subsection{Crop Damage: The Primary Barrier}
\label{sec:cropdamage}
Crop contact rates of 20\nobreakdash--29\% at all tested speeds represent a critical barrier to practical deployment, exceeding the established 10\% tolerance threshold~\cite{slaughter2008} by approximately 2\nobreakdash--3$\times$. Three mechanisms explain this elevated crop contact. First, spatial proximity constrained targeting precision: weeds within 2\nobreakdash--5~cm of crop plants fell within the positional error envelope of the present solenoid-actuated, 12-nozzle sprayer given the 2.5~cm in-row radish spacing used here. Second, morphological similarity between young radish and weeds (Palmer amaranth) likely contributed to classification errors that the IoU $> 0.25$ crop overlap filter could not fully suppress. Third, speed-dependent timing errors increasingly displaced spray toward adjacent crop plants at higher platform speeds. Together, these mechanisms reflect limitations in the spatial and temporal precision of the present actuation pipeline rather than an absolute physical limit on crop-adjacent targeting.

A recently reported sprayer using the same 12-nozzle, 1~cm nozzle-pitch geometry as the present system, but with micro-jet rather than solenoid actuation, achieved a lower crop hit rate of 17.6\% in a lettuce field at a similar platform speed ($\sim$0.25~m/s)~\cite{deng2026b}, indicating that actuator technology, not nozzle spacing, is a meaningful lever for reducing crop contact. Notably, however, even this higher-precision sprayer remained above the 10\% commercial tolerance. This result is consistent with Deng \textit{et al.}~\cite{deng2026b}, who concluded that further reducing crop contact will likely require more robust weed detection and an explicit buffer zone around crops, in addition to actuator improvements. Addressing this barrier in the present system therefore requires a coordinated hardware-software strategy incorporating explicit weed--crop distance thresholding, faster valve actuators or alternative actuation technologies, and appearance-based association within the existing tracking pipeline (e.g., incorporating visual feature similarity alongside the current IoU-based matching in ByteTrack) to reduce identity switches between morphologically similar species.

These results characterize the present system's crop-adjacent targeting precision rather than offering a general verdict on vision-guided precision spraying relative to other weed control strategies; a full economic and performance comparison against broadcast herbicide, mechanical cultivation, and laser weeding is beyond the scope of the single-platform, single-crop trials reported here. What this study does establish, corroborated by the independent, higher-precision system discussed above, is that spray-timing latency and actuator resolution, not detection accuracy, are the binding constraints on crop-adjacent targeting across at least two independent system generations in this research program, and that closing them is a well-defined engineering target for near-term development, discussed next.

Several limitations constrain the generalizability of these findings and define important directions for future work. Testing was conducted at a single site on one crop species under favorable conditions (flat terrain, minimal wind, overcast sky) with two dominant weed species. These conditions may limit extrapolation to the broader diversity of vegetable production environments. The tested speed range of 0.13\nobreakdash--0.45~m/s did not reach the 0.67~m/s commercial threshold, requiring extrapolation that likely underestimates actual degradation given the non-linear acceleration of performance losses observed between 0.27 and 0.45~m/s. The spray-timing model's valve-response and droplet-flight latency terms (Section~\ref{sec:timing}) were based on nominal values rather than bench-validated measurements for the hardware used in this study; only the software-processing component was directly measured (Table~\ref{tab:yolo}). All three field speed treatments were tested sequentially, single-pass, on the same experimental rows rather than on independent, replicated plots. Although the approximately 3\nobreakdash--4~h interval between speeds was not expected to confound ground-truth scoring, dye carryover and row-order effects were not independently verified. The use of water-dye markers provided a proxy for spray deposition rather than herbicide efficacy, leaving actual weed mortality unvalidated.

Further research should therefore prioritize extending speed characterization to and beyond commercial operating thresholds, conducting multi-crop validations, and improving motion-robust perception through video object detection with temporal attention. Integration of RTK-GPS velocity measurements could further improve timing compensation across variable platform speeds. Additional work is still needed to bench-validate valve response and droplet-flight latency, evaluate faster valve actuators or alternative actuation technologies, and conduct herbicide efficacy trials that directly quantify weed mortality. Longer-term research should include multi-site validation, economic comparisons with conventional and alternative weeding methods, adaptive speed control based on real-time weed density, and continued actuator-technology development focused specifically on overcoming the crop-adjacent contact limitation identified here as the primary practical barrier.

\section{Conclusion}
\label{sec:conclusion}
This multi-speed field evaluation demonstrates that spraying accuracy deteriorated more rapidly than detection accuracy as platform speed increased. The weed hit rate decreased by 43.3 percentage points, compared with a 30.7-point decline in detection recall, indicating that downstream tracking and actuation, rather than plant perception, were the primary constraints on spraying performance. Crop contact, which remained between 20\nobreakdash--29\% across all tested speeds, represents an additional critical barrier, exceeding the 10\% tolerance threshold regardless of operating speed. The system met none of the three commercial-viability criteria ($>$90\% weed control, $<$10\% crop damage, $\geq 0.67$~m/s operating speed), even at its best-performing, slowest tested speed. These findings identify three key engineering priorities: motion-robust detection, spray-timing compensation, and improved crop-adjacent actuation precision. Addressing these limitations is essential for advancing smart spraying in vegetable production.

\section*{Acknowledgments}
This work was supported by the Farm Innovation Grant from the Michigan Department of Agriculture \& Rural Development. The authors gratefully acknowledge the Michigan State University Horticulture Teaching and Research Center for providing field access and technical support.

\section*{Declarations}
\textit{Author contributions (CRediT).} Md Khairul Islam: Conceptualization, Methodology, Software, Investigation, Data curation, Formal analysis, Visualization, Writing (original draft). Boyang Deng: Investigation, Data curation, Methodology. Yuzhen Lu: Conceptualization, Methodology, Supervision, Project administration, Funding acquisition, Writing (review \& editing). Daniel Brainard: Resources, Investigation. Ajit Srivastava: Resources, Investigation.

\textit{Competing interests.} The authors declare that they have no known competing financial interests or personal relationships that could have appeared to influence the work reported in this paper.

\textit{Data and code availability.} The annotated dataset (Weed\&CropDet2025 component) and trained model weights will be made available upon reasonable request to the corresponding author. The previously published 3SeasonWeedDet10 dataset is available from Deng and Lu~\cite{denglu2025}. Code for the detection, tracking, and spray-timing control pipeline is available at \url{https://github.com/AgFood-Sensing-and-Intelligence/multispeed-precision-sprayer}.

\textit{Supplementary material.} Supplementary Table S1: complete hardware specifications and navigation retrofit cost breakdown (Sections~\ref{sec:arch}--\ref{sec:actuator}). Supplementary Table S2: complete training hyperparameters for the YOLOv11 detection model (Section~\ref{sec:detmodel}). Supplementary Table S3: complete failure mode taxonomy for unsuccessful spray events (Section~\ref{sec:analytical}). Supplementary Video S1: field detection, tracking, and live spray actuation recorded during the multi-speed trials (August 29--September 4, 2025), illustrating the targeting behavior underlying the results in Table~\ref{tab:sprayspeed}.


\bibliographystyle{IEEEtran}
\bibliography{Reference}
\end{document}